\documentclass[12pt]{article} 
\usepackage{amsmath,amsthm,amssymb,bm,mathrsfs,}
\usepackage{algorithm,algorithmic,graphicx,subfig}
\usepackage{caption,enumerate,natbib,listings,setspace}
\usepackage[OT1]{fontenc}
\usepackage[colorlinks,linkcolor=red,anchorcolor=blue,citecolor=blue]{hyperref}
\usepackage[margin=1in]{geometry}
\usepackage[protrusion=false,expansion=true]{microtype}
\usepackage[title]{appendix}
\usepackage{bbm}
\theoremstyle{plain}

\let\hat\widehat
\let\tilde\widetilde

\newcommand{\ab}{\mathbf{a}}

\newcommand{\gb}{\mathbf{g}}

\newcommand{\vb}{\mathbf{v}}

\newcommand{\xb}{\mathbf{x}}

\newcommand{\Ab}{\mathbf{A}}
\newcommand{\Bb}{\mathbf{B}}

\newcommand{\Db}{\mathbf{D}}

\newcommand{\Mb}{\mathbf{M}}

\newcommand{\Ub}{\mathbf{U}}

\newcommand{\cA}{\mathcal{A}}
\newcommand{\cB}{\mathcal{B}}
\newcommand{\cC}{\mathcal{C}}

\newcommand{\cE}{\mathcal{E}}
\newcommand{\cF}{\mathcal{F}}

\newcommand{\cR}{\mathcal{R}}
\newcommand{\cS}{{\mathcal{S}}}
\newcommand{\cT}{{\mathcal{T}}}

\newcommand{\cY}{\mathcal{Y}}

\newcommand{\RR}{\mathbb{R}}

\newcommand{\bbeta}{\bm{\beta}}

\newcommand{\bLambda}{\bm{\Lambda}}

\newcommand{\argmin}{\mathop{\mathrm{argmin}}}

\DeclareMathOperator{\ind}{\mathrm{1}}  

\newtheorem{proposition}{{\bf Proposition}}

\newtheorem{theorem}{{\bf Theorem}}

\newtheorem{assumption}{{\bf Assumption}}

\newtheorem{definition}{{\bf Definition}}

\newcommand{\change}[1]{{\leavevmode\color{black}{#1}}}

\title{\Large{\textbf{Partially Observed Dynamic Tensor \\}} 
\Large{\textbf{Response Regression}}} 

\author
{
Jie Zhou\thanks{PhD student, Department of Management Science,  University of Miami Herbert Business School, Miami, FL 33146. Email: jzhou@bus.miami.edu.},
Will Wei Sun\thanks{Assistant Professor, Krannert School of Management, Purdue University, IN 47907. Email: sun244@purdue.edu.}, 
Jingfei Zhang\thanks{Associate Professor, Department of Management Science, University of Miami Herbert Business School, Miami, FL 33146. Email: ezhang@bus.miami.edu.}, and
Lexin Li\thanks{Professor, Division of Biostatistics, University of California, Berkeley, Berkeley, CA 94720. Email: lexinli@berkeley.edu.}
}

\date{}
\begin{document} 

\maketitle

\begin{abstract}
\noindent
In modern data science, dynamic tensor data is prevailing in numerous applications. An important task is to characterize the relationship between such dynamic tensor and external covariates. However, the tensor data is often only partially observed, rendering many existing methods inapplicable. In this article, we develop a regression model with partially observed dynamic tensor as the response and external covariates as the predictor. We introduce the low-rank, sparsity and fusion structures on the regression coefficient tensor, and consider a loss function projected over the observed entries. We develop an efficient non-convex alternating updating algorithm, and derive the finite-sample error bound of the actual estimator from each step of our optimization algorithm. Unobserved entries in tensor response have imposed serious challenges. As a result, our proposal differs considerably in terms of estimation algorithm, regularity conditions, as well as theoretical properties, compared to the existing tensor completion or tensor response regression solutions. We illustrate the efficacy of our proposed method using simulations, and two real applications, a neuroimaging dementia study and a digital advertising study. 
\end{abstract}


\newpage
\baselineskip=21pt

\section{Introduction}
\label{sec:introduction}

In modern data science, dynamic tensor data is becoming ubiquitous in a wide variety of scientific and business applications. The data takes the form of a multidimensional array, and one mode of the array is time, giving the name dynamic tensor. It is often of keen interest to characterize the relationship between such time-varying tensor data and external covariates. One example is a neuroimaging study of Alzheimer's disease (AD) \citep{thung2016}. Anatomical magnetic resonance imaging (MRI) data are collected for 365 individuals with and without AD every six month over a two-year period. Each image, after preprocessing, is of dimension $32 \times 32 \times 32$, and the combined data is in the form of a subject by MRI image by time tensor. An important scientific question is to understand how a patient's structural brain atrophy is associated with clinical and demographic characteristics such as the patient's diagnosis status, age and sex. Another example is a digital advertising study \citep{bruce2017dynamic}. The click-through rate (CTR) of 20 active users reacting to digital advertisements from 2 publishers are recorded for 80 advertisement campaigns on a daily basis over a four-week period. The data is formed as a tensor of campaign by user by publisher by time. An important business question is to understand how features of an advertisement campaign affect its effectiveness measured by CTR on the target audience. Both questions can be formulated as a supervised tensor learning problem. However, a crucial but often overlooked issue is that the data is likely only partially observed in real applications. For instance, in the neuroimaging study, not all individuals have completed all five biannual MRI scans in two years. In the digital advertising study, not all users are exposed to all campaigns, nor react to all publishers. Actually, in our digital advertising data, more than 95\% of the entire tensor entries are unobserved. In this article, we tackle the problem of supervised tensor learning with partially observed tensor data.  

There are several lines of research that are closely related to but also clearly distinctive of the problem we address. The first line studies tensor completion \citep{JO2014, yuan2016on, yuan2017incoherent, dong2017, zhang2019cross}. Tensor completion aims to fill in the unobserved entries of a partially observed tensor, usually by resorting to some tensor low-rank and sparsity structures. It is unsupervised learning, as it involves no external covariates. While we also tackle tensor with unobserved entries and we are to employ similar low-dimensional structures as tensor completion, our goal is not to complete the tensor. Instead, we target a supervised learning problem, and aim to estimate the relationship between the partially observed tensor and external covariates. Consequently, our model formulation, estimation approach, and theoretical analysis are considerably different from tensor completion. The second line tackles tensor regression where the response is a scalar and the predictor is a tensor \citep{zhou2013, WangZhu2016, hzc20, hwz20}. By contrast, we treat tensor as the response and covariates as the predictor. When it comes to theoretical analysis, the two models involve utterly different techniques. The third line studies regressions with a tensor-valued response, while imposing different structural assumptions on the resulting tensor regression coefficient \citep{rabusseau2016, li2016, store2017,ChenYuan2019, xhw19}. This line of works share a similar goal as ours; however, none of these existing methods can handle tensor response with partially observed entries. Moreover, none is able to pool information from the dynamic tensor data collected at adjacent time points. In our experiments, we show that, focusing only on the subset of completely observed tensor data, or ignoring the structural smoothness over time  would both lead to considerable loss in estimation accuracy. Finally, there have been a number of proposals motivated by similar applications and can handle missing values. Particularly, \cite{magee} considered an adaptive voxel-wise approach by modeling each entry of the dynamic tensor seperately. We instead adopt a tensor regression approach by jointly modeling all the entries of the entire tensor. We later numerically compare our method with \cite{magee} and other solutions. \citet{XueQu2019} studied regressions of multi-source data with missing values involving neuroimaging features. However, the images were summarized as a vector instead of a tensor, and were placed on the predictor side. Similarly, \citet{feng2019bayesian} developed a scalar-on-image regression model with missing image scans. By contrast, we place the imaging tensor on the response side. 

In this article, we develop a regression model with partially observed dynamic tensor as the response. We impose that the coefficient tensor to be both sparse and low-rank, which reduces the dimension of the parameter space, lessens the computational complexity, and improves the interpretability of the model. Furthermore, we impose a fusion structure along the temporal mode of the tensor coefficient, which helps pool the information from data observed at adjacent time points. All these assumptions are scientifically plausible, and have been widely used in numerous applications including both neuroimaging analysis and digital marketing \citep{ vounou2010discovering, zhou2013, yin2015dynamic, rabusseau2016,bi2018multilayer, tang2019individualized, zhang2019tensor}. To handle the unobserved entries of the tensor response, we consider a loss function projected over the observed entries, which is then optimized under the low-rank, sparsity and fusion constraints. We develop an efficient non-convex alternating updating algorithm, and derive the finite-sample error bound of the actual estimator from each step of our optimization algorithm. 

Unobserved entries in tensor response have introduced serious challenges, as the existing algorithms for estimating a sparse low-rank tensor and technical tools for asymptotic analysis are only applicable to either a single partially observed tensor or a fully observed tensor \citep[e.g.,][]{JO2014, store2017}. As a result, our proposal differs considerably in terms of estimation algorithm, regularity conditions, as well as theoretical properties. For estimation, since the unobserved entries can occur at different locations for different tensors, the loss function projected over the observed entries takes a complex form. The traditional vector-wise updating algorithms \citep{JO2014,store2017} are no longer applicable. Alternatively, we propose a new procedure that updates the low-rank components of the coefficient tensor in an element-wise fashion; see Step 1 of Algorithm \ref{alg:STD} and equation \eqref{eqn:betanew} in Section \ref{sec:estimation}. For regularity conditions, we add a $\mu$-mass condition to ensure that  sufficient information is contained in the observed entries for tensor coefficient estimation; see Assumption \ref{ass:model}. We also place a lower bound on the probability of the observation $p$, and discuss its relation with the sample size, tensor dimension, sparsity level and mass parameter $\mu$; see Assumptions \ref{ass:missing} and \ref{ass:missing_r}. Our lower bound is different from that in the tensor completion literature \citep{JO2014, yuan2016on, yuan2017incoherent, dong2017}, which considered only a single tensor, whereas we consider a collection of $n$ tensors. Consequently, our lower bound on $p$ depends on $n$, and tends to $0$ as $n$ tends to infinity. For theoretical properties, we show that the statistical error of our estimator has an interesting connection with the lower bound on $p$, which does not appear in the tensor response regression for complete data \citep{store2017}. This characterizes the loss at the statistical level when modeling with only partially observed tensor. In summary, our proposal is far from an incremental extension from the complete case scenario, and involves a new set of strategies for estimation and theoretical analysis. 

We adopt the following notation throughout the article. Let $[d] = \{1,\ldots,d\}$, and let $\circ$ and $\otimes$ denote the outer product and kronecker product. For a vector $\ab \in \RR^d$, let $\|\ab\|$ and $\|\ab\|_0$ denote its Euclidean norm and $\ell_0$ norm, respectively. For a matrix $\Ab \in \RR^{d_1\times d_2}$, let $\|\Ab\|$ denote its spectral norm. For a tensor $\cA \in \RR^{d_1\times \ldots \times d_m}$, let $\cA_{i_1,\cdots,i_m}$ be its $(i_1,\cdots, i_m)$th entry, and $\cA_{i_1,\cdots,i_{j-1},:,i_{j+1},\ldots,i_m}=(\cA_{i_1,\cdots,i_{j-1},1,i_{j+1},\ldots,i_m},\ldots,\cA_{i_1,\cdots,i_{j-1},d_j,i_{j+1},\ldots,i_m})^{\top}\in\mathbb{R}^{d_j}$. Let $\text{unfold}_m(\cA)$ denote the mode-$m$ unfolding of $\cA$, which arranges the mode-$m$ fibers to be the columns of the resulting matrix; e.g., the mold-1 unfolding of a third-order tensor $\cA\in\mathbb{R}^{d_1\times d_2\times d_3}$ is $\text{unfold}_1(\cA)=[\cA_{:,1,1},\ldots,\cA_{:,d_2,1},\ldots,\cA_{:,d_2,d_3}]\in\mathbb{R}^{d_1\times(d_2d_3)}$. Define the tensor spectral norm as $\| \cA \| = \sup_{\|\ab_1\|=\ldots=\|\ab_m\|=1} | \cA \times_1 \ab_1 \times_2 \ldots  \times_m \ab_m|$, and the tensor Frobenius norm as $\| \cA \|_F= \sqrt{ \sum_{i_1,\ldots,i_m} \cA_{i_1,\ldots,i_m}^2}$. For $\ab \in \RR^{d_j}$, define the $j$-mode tensor product as $\cA \times_{j} \ab \in \RR^{d_1\times\cdots\times d_{j-1}\times d_{j+1}\times\cdots\times d_m}$, such that $(\cA \times_j \ab)_{i_1,\cdots, i_{j-1},i_{j+1},\cdots, i_m} = \sum_{i_j=1}^{d_j}\cA_{i_1,\cdots, i_m} a_{i_j}$. For $\ab_j \in \RR^{d_j}, j\in[m]$, define the multilinear combination of the tensor entries as $\cA \times_1 \ab_1 \times_2 \ldots \times_m \ab_m = \sum_{i_1\in[d_1]} \ldots \sum_{i_m\in [d_m]} a_{1,i_1} \ldots a_{m,i_m} \cA_{i_1,\ldots,i_m}$, where $a_{j,i_j}$ is the $i_j$th entry of $\ab_{j}$. For two sequences $a_n, b_n$, we say $a_n = \mathcal{O}(b_n)$ if $a_n \le C b_n$ for some positive constant $C$.

The rest of the article is organized as follows. Section \ref{sec:model} introduces our regression model with partially observed dynamic tensor response. Section \ref{sec:estimation} develops the estimation algorithm. Section \ref{sec:theorem} investigates the theoretical properties. Section \ref{sec:simulations} presents the simulation results, and Section \ref{sec:realdata} illustrates with two real world datasets, a neuroimaging study and a digital advertising study.  All technical proofs are relegated to the Supplementary Materials.

\section{Model}
\label{sec:model}

Suppose at each time point $t$, we collect an $m$th-order tensor $\cY_t$ of dimension $d_1\times \ldots \times d_m$, $t \in [T]$. We stack the collected tensors $\cY_1,\ldots,\cY_T$ together, and represent it as an $(m+1)$th-order tensor $\cY \in \mathbb{R}^{d_1 \times \cdots \times d_m\times T}$. Correspondingly, the $(m+1)$th mode of $\cY$ is referred as the temporal mode. Suppose there are totally $n$ subjects in the study. For each subject $i$, we collect a dynamic tensor represented as $\cY_i$, along with a $q$-dimensional vector of covariates $\xb_i \in \mathbb{R}^{q}$, $i \in [n]$. The response tensor $\cY_i$ can be partially observed, and the missing patterns can vary from subject to subject. We consider the following regression model,
\begin{equation} \label{eqn:true_model}
\cY_i = \cB ^*\times_{m+2} \xb_i + \cE_i,
\end{equation}
where $\cB^* \in \mathbb{R}^{d_1 \times \cdots \times d_m \times T\times q}$ is an $(m+2)$th-order coefficient tensor, and $\cE_i \in \mathbb{R}^{d_1 \times \cdots \times d_m\times T}$ is an $(m+1)$th-order error tensor independent of $\xb_i$. Without loss of generality, we assume the data are centered, and thus drop the intercept term in model \eqref{eqn:true_model}. The coefficient tensor $\cB^*$ captures the relationship between the dynamic tensor response and the predictor, and is the main object of interest in our analysis. For instance, $\cB^*_{i_1,\cdots,i_m,:,l}\in\mathbb{R}^T$ describes the effect of the $l$th covariate on the time-varying pattern of the $(i_1,\ldots,i_m)$th entry of tensor $\cY_t$. Next, we impose three structures on $\cB^*$ to facilitate its analysis. 

We first assume that $\cB^*$ admits a rank-$r$ CP decomposition structure, in that,  
\begin{equation} \label{eqn:sparse_tensor}
\cB^* = \sum_{k\in [r]} w^*_k \bbeta^*_{k,1} \circ \cdots \circ \bbeta^*_{k, m+2},
\end{equation}
where $\bbeta^*_{k,j} \in \mathbb{S}^{d_j}$, $\mathbb{S}^{d} = \{\ab \in \mathbb{R}^d \,|\, \|\ab\| = 1\}$, and $w^*_k>0$. The CP structure is one of the most common low-rank structures \citep{kolda2009tensor}, and is widely used in tensor data analysis \citep[among others]{zhou2013, anandkumar2014a, JO2014, yuan2016on, yuan2017incoherent, zhang2019cross, ChenYuan2019}. We next assume that $\cB^*$ is sparse, in that the decomposed components $\bbeta^*_{k,j}$'s are sparse. That is, $\bbeta_{k,j}^{*}\in \cS(d_j,s_{j})$ for $j \in [m+1], k \in [r]$, where 
\vspace{-0.05in}
\begin{eqnarray*}
\cS(d, s) = \left\{ \bbeta \in \mathbb R^d \mid \sum_{l=1}^{d} \ind_{(\beta_{l}  \ne 0)} \le s \right\} = \left\{ \bbeta \in \mathbb R^{d} \mid  \left \|  \bbeta \right \|_0 \le s \right\}.
\end{eqnarray*}
This assumption postulates that the covariates $\xb$'s effects are concentrated on a subset of entries of $\cB^*$, which enables us to identify most relevant regions in the dynamic tensor that are affected by the covariates. The sparsity assumption is again widely employed in numerous applications including neuroscience and online advertising \citep{bullmore2009complex, vounou2010discovering, sun2017provable}. We further assume a fusion structure on the decomposed components $\bbeta^*_{k,j}$ of $\cB^*$. That is, $\bbeta_{k,j}^{*} \in \cF(d_j,f_{j})$ for $j \in [m+1], k \in [r]$, where 
\begin{eqnarray*}
\cF(d, f) = \left\{ \bbeta \in \mathbb R^{d} \mid  \sum_{l=2}^{d}\ind_{( \left |\beta_{l} - \beta_{l-1} \right|\ne 0 )} \le f \right\} = \left\{ \bbeta \in \mathbb R^{d} \mid  \left \|  \Db \bbeta \right \|_0 \le f-1 \right\},
\end{eqnarray*}
and $\Db \in \mathbb R^{(d-1) \times d}$ with $\Db_{i, i}=-1$, $\Db_{i, i+1}=1$ for $i \in [d-1]$, and other entries being zeros. This assumption encourages temporal smoothness and helps pool information from tensors observed at adjacent time points \citep{madrid2017, SunLi2019}. Putting the sparsity and fusion structures together, we have
\begin{equation} \label{eqn:parspace}
\bbeta_{k,j}^{*}\in\cS(d_j,s_j)\cap\cF(d_j,f_j),\quad\text{for } j\in [m+1],\,k\in [r].
\end{equation}
We briefly comment that, since the dimension $q$ of the covariates $\xb$ is relatively small in our motivating examples, we have chosen not to impose any sparsity or fusion structure on the component $\bbeta_{k,m+2}^{*} \in\mathbb{R}^q$, which is the last mode of the coefficient tensor $\cB^*$. Nevertheless, we can easily incorporate such a structure for $\bbeta_{k,m+2}^{*}$, or other structures. The extension is straightforward, and thus is not further pursued. 

A major challenge we face is that many entries of the dynamic tensor response $\cY$ are unobserved. Let $\Omega\subseteq[d_1]\times [d_2]\times\cdots\times[d_{m+1}]$ denote the set of indexes for the observed entries, and $\Omega_i$ denote the set of indexes for the observed entries in $\cY_i$, $i \in [n]$. We define a projection function $\Pi_\Omega(\cdot)$ that projects the tensor onto the observed set $\Omega$, such that 
\begin{equation*}
[\Pi_{\Omega}(\cY)]_{i_1,i_2,\dots, i_{m+1}}=\begin{cases}
\cY_{i_1,i_2,\dots, i_{m+1}} &\mbox{if $(i_1,\dots ,i_{m+1})\in \Omega$,}\\
0&\mbox{otherwise}.
\end{cases}
\end{equation*}
We then consider the following constrained optimization problem,
\begin{align} \label{eqn:objective}
& \min_{\substack{w_k, \bbeta_{k,j} \\ k \in [r],  j \in [m+2]}} \; 
\frac{1}{n}\sum_{i = 1}^n \left\| \Pi_{\Omega_i}\left( \cY_i - \sum_{k\in [r]} w_k (\bbeta_{k,m+2}^{\top} \xb_i) \bbeta_{k,1} \circ \cdots \circ \bbeta_{k,m+1} \right)  \right\|_F^2  \\ 
& \textrm{ subject to } \|\bbeta_{k,j}\|_2 = 1, j\in[m+2], \|\bbeta_{k,j}\|_0 \le \tau_{s_{j}},  \left \|  \Db \bbeta_{k,j} \right \|_0 \le \tau_{f_{j}}, j\in[m+1], k\in [r]. \nonumber
\end{align}
In this optimization, both sparsity and fusion structures are imposed through $\ell_0$ penalties. Such non-convex penalties have been found effective in high-dimensional sparse models \citep{shen2012, Zhu2014} and fused sparse models \citep{rinaldo2009, wang2016}.

\section{Estimation}
\label{sec:estimation}

The optimization problem in \eqref{eqn:objective} is highly nontrivial, as it is a non-convex optimization with multiple constraints and a complex loss function due to the unobserved entries. We develop an alternating block updating algorithm to solve \eqref{eqn:objective}, and divide our procedure into multiple alternating steps. First, we solve an unconstrained weighted tensor completion problem, by updating $\bbeta_{k,1}, \ldots, \bbeta_{k,m+1}$, given $w_k$ and $\bbeta_{k,m+2}$, for $k\in[r]$. Since each response tensor is only partially observed and different tensors may have different missing patterns, the commonly used vector-wise updating approach in tensor analysis is no longer applicable. To address this issue, we propose a new element-wise approach to update the decomposed components of the low-rank tensor. Next, we define a series of operators and apply them to the unconstrained estimators obtained from the first step, so to incorporate the sparsity and fusion constraints on $\bbeta_{k,1}, \ldots, \bbeta_{k,m+1}$. Finally, we update $w_k$ and $\bbeta_{k,m+2}$, both of which have closed-form solutions. We summarize the procedure in Algorithm \ref{alg:STD}, then discuss each step.

\begin{algorithm}[t!]
\caption{Alternating block updating algorithm for \eqref{eqn:objective}}
\begin{algorithmic}[1]
\STATE \textbf{input:}  the data $\big\{ (\xb_i,\cY_i,\Omega_i), i=1, \ldots, n \big\}$, the rank $r$, the sparsity parameter $\tau_{s_j}$, and the fusion parameter $\tau_{f_j}$, $j \in [m+1]$.
\STATE \textbf{initialization:} set $w_k=1$, and randomly generate unit-norm vectors $\bbeta_{k,1},\ldots, \bbeta_{k,m+2}$ from a standard normal distribution, $k\in [r]$.
\REPEAT
\FOR{$k=1$ to $r$}
\FOR{$j=1$ to $m+1$}
\STATE step 1: obtain the unconstrained estimator $\tilde{\bbeta}^{(t+1)}_{k,j}$, \change{given $\hat{w}^{(t)}_k$, $\hat{\bbeta}^{(t+1)}_{k,1},\ldots,\hat{\bbeta}^{(t+1)}_{k,j-1}$, $\hat{\bbeta}^{(t)}_{k,j+1}, \ldots, \hat{\bbeta}^{(t)}_{k,m+1}, \hat{\bbeta}^{(t)}_{k,m+2}$,} by solving \eqref{eqn:step1}; normalize $\tilde{\bbeta}^{(t+1)}_{k,j}$. 
\STATE step 2: obtain the constrained estimator $\hat{\bbeta}^{(t+1)}_{k,j}$, by applying the \texttt{Truncatefuse} operator to $\tilde{\bbeta}^{(t+1)}_{k,j}$; normalize $\hat{\bbeta}^{(t+1)}_{k,j}$.  
\ENDFOR 
\STATE step 3: obtain $\hat{w}^{(t+1)}_k$, given $\hat{\bbeta}^{(t+1)}_{k,1},\ldots,\hat{\bbeta}^{(t+1)}_{k,m+1}, \hat{\bbeta}^{(t)}_{k,m+2}$, using \eqref{eqn:what}. 
\STATE step 4: obtain $\hat{\bbeta}^{(t+1)}_{k,m+2}$, given $\hat{w}^{(t+1)}_k, \hat{\bbeta}^{(t+1)}_{k,1},\ldots,\hat{\bbeta}^{(t+1)}_{k,m+1}$, using \eqref{eqn:betam2}. 
\ENDFOR 
\UNTIL the stopping criterion is met. 
\STATE\textbf{output:} $\hat{w}_k, \widehat{\bbeta}_{k,1},\ldots,\widehat{\bbeta}_{k,m+2}$, $k \in [r]$.
\end{algorithmic}
\label{alg:STD}
\end{algorithm}

In step 1, we solve an unconstrained weighted tensor completion problem, \change{
\begin{equation} \label{eqn:step1}
\min\limits_{\bbeta_{k,j}} \; \frac{1}{n} \sum_{i = 1}^n \left\{ \alpha_{i,k}^{(t)} \right\}^2 \left\| \Pi_{\Omega_i} \left( \cR_{i,k}^{(t+1)} - \hat{w}^{(t)}_k \hat{\bbeta}^{(t+1)}_{k,1} \circ \cdots \circ \hat{\bbeta}^{(t+1)}_{k,j-1} \circ \bbeta_{k,j} \circ \hat{\bbeta}^{(t)}_{k,j+1}\circ  \cdots \circ \hat{\bbeta}^{(t)}_{k,m+1} \right) \right\|_F^2, 
\end{equation}}
where $\alpha_{i,k}^{(t)} = \bbeta_{k,m+2}^{(t)\top}\xb_i$, and $\cR_{i,k}^{(t+1)}$ is a residual term defined as, \change{
\begin{equation} \label{eqn:resid}
\cR_{i,k}^{(t+1)} = \left( \cY_i -\sum_{{k'} < k} \hat{w}_{k'}^{(t+1)} \alpha_{i,k'}^{(t+1)} \bbeta_{k',1}^{(t+1)} \circ \ldots \circ \bbeta_{k',m+1}^{(t+1)} -\sum_{{k'} > k} \hat{w}_{k'}^{(t)} \alpha_{i,k'}^{(t)} \bbeta_{k',1}^{(t)} \circ \ldots \circ \bbeta_{k',m+1}^{(t)} \right) / \alpha_{i,k}^{(t)},
\end{equation}}
for $i \in [n], k \in [r]$. The optimization problem in \eqref{eqn:step1} has a closed-form solution. To simplify the presentation, we give this explicit expression when $m=2$. For the case of $m \ge 3$, the calculation is similar except involving more terms. Specifically, the $l$th entry of $\tilde{\bbeta}_{k,3}^{(t+1)}$ is 
\begin{equation} \label{eqn:betanew}
\tilde{\bbeta}^{(t+1)}_{k,3,l} = \frac{\sum_{i=1}^{n} \left\{ \alpha_{i,k}^{(t)} \right\}^2 \sum_{l_1,l_{2}}\delta_{i,l_1,l_2,l}\, \cR_{i,k,l_1,l_2,l}^{(t+1)}\,\hat{\bbeta}^{(t+1)}_{k,1,l_1} \hat{\bbeta}^{(t+1)}_{k,2,l_2}}{\sum_{i=1}^{n} \left\{ \alpha_{i,k}^{(t)} \right\}^2 \sum_{l_1,l_2} \hat{w}_k^{(t)} \delta_{i,l_1,l_2,l}\, \left\{\hat{\bbeta}^{(t+1)}_{k,1,l_1} \right\}^2  \left\{ \hat{\bbeta}^{(t+1)}_{k,2,l_2} \right\}^2}, 
\end{equation}
where $\delta_{i,l_1,l_2,l} = 1$ if $(l_1,l_2,l)\in \Omega_i$, and $\delta_{i,l_1,l_2,l} = 0$ otherwise. Here $\cR^{(t+1)}_{i,k,l_1,l_2,l}$ refers to the $(l_1,l_2,l)$th entry of $\cR^{(t+1)}_{i,k}$.
The expressions for $\tilde{\bbeta}_{k,1}^{(t+1)}$ and $\tilde{\bbeta}_{k,2}^{(t+1)}$ can be derived similarly. We remark that, \eqref{eqn:betanew} is the key difference between our estimation method and those for a single partially observed tensor \citep{JO2014}, or a completely observed tensor \citep{store2017}. Particularly, the observed entry indicator $\delta_{i,l_1,l_2,l}$ appears in both the numerator and denominator, and $\delta_{i,l_1,l_2,l}$ is different across different entries of $\tilde{\bbeta}^{(t+1)}_{k,3}$. Therefore, $\tilde{\bbeta}^{(t+1)}_{k,3}$ needs to be updated in an element-wise fashion, as $\delta_{i,l_1,l_2,l}$ could not be cancelled. After obtaining \eqref{eqn:betanew}, we normalize $\tilde{\bbeta}^{(t+1)}_{k,j}$ to ensure a unit norm. 

In step 2, we apply the sparsity and fusion constraints to $\tilde{\bbeta}^{(t+1)}_{k,j}$ obtained in the first step. Toward that goal, we define a truncation operator \texttt{Truncate}($\ab, \tau_s$), and a fusion operator \texttt{Fuse}($\ab, \tau_f$), for a vector $\ab \in\mathbb{R}^{d}$ and two integer-valued tuning parameters $\tau_s$ and $\tau_f$, as, 
\begin{eqnarray*}
[\texttt{Truncate}(\ab, \tau_s)]_{j} =   
\begin{cases}
a_{j} & \mbox{if $j \in \text{supp}(\ab ,\tau_s)$} \\
0 & \mbox{otherwise}
\end{cases};  \quad\quad
[\texttt{Fuse}(\ab, \tau_f)]_j  =  \sum_{i=1}^{\tau_f} \ind_{j\in \cC_i} \frac{1}{|\cC_i|} \sum_{l \in \cC_i} a_l,
\end{eqnarray*}
where supp($\ab,\tau_s$) refers to the indexes of $\tau_s$ entries with the largest absolute values in $\ab$, and $\{\cC_i\}_{i=1}^{\tau_f}$ are the fusion groups. This truncation operator ensures that the total number of nonzero entries in $\ab$ is bounded by $\tau_s$, and is commonly employed in non-convex sparse optimizations \citep{yuan2013, sun2017provable}. The fusion groups $\{\cC_i\}_{i=1}^{\tau_f}$ are calculated as follows. First, the truncation operator is applied to $\Db\ab\in \RR^{d-1}$. The resulting $\texttt{Truncate}(\Db \ab, \tau_f-1)$ has at most $(\tau_f-1)$ nonzero entries. Then the elements $a_j$ and $a_{j+1}$ are put into the same group if $[\texttt{Truncate}(\Db \ab, \tau_f-1)]_j = 0$. This procedure in effect groups the elements in $\ab$ into $\tau_f$ distinct groups, which we denote as $\{\cC_i\}_{i=1}^{\tau_f}$. Elements in each of the $\tau_f$ groups are then averaged to obtain the final result. Combining the two operators, we obtain the $\texttt{Truncatefuse}(\ab, \tau_s, \tau_f)$ operator as, 
\begin{equation*} \label{eqn:tf}
\texttt{Truncatefuse}(\ab, \tau_s, \tau_f) = \texttt{Truncate}\big\{ \texttt{Fuse}(\ab, \tau_f), \tau_s \big\},
\end{equation*}
where $\tau_s \le d$ is the sparsity parameter, and $\tau_f \le d$ is the fusion parameter. For example, consider $\ab = (0.1, 0.2, 0.4, 0.5, 0.6)^{\top}$, $\tau_s = 3$ and $\tau_f = 2$. Correspondingly, $\Db\ab = (0.1,0.2,0.1,0.1)^{\top}$. We then have $\texttt{Truncate}(\Db\ab,\tau_f-1) = (0, 0.2,0,0)^{\top}$. This in effect suggests that $a_1, a_2$ belong to one group, and $a_3, a_4, a_5$ belong to the other group. We then average the values of $\ab$ in each group, and obtain $\texttt{Fuse}(\ab, \tau_f) = (0.15,0.15,0.5,0.5,0.5)^{\top}$. Lastly, $\texttt{Truncatefuse}(\ab, \tau_s, \tau_f) = \texttt{Truncate}\big\{ \texttt{Fuse}(\ab, \tau_f), \tau_s \big\} = \texttt{Truncate}\big\{ (0.15,0.15,0.5,0.5,0.5)^{\top},$ $3 \big\} = (0,0,0.5,0.5,0.5)^{\top}$. We apply the \texttt{Truncatefuse} operator to the unconstrained estimator $\tilde{\bbeta}^{(t+1)}_{k,j}$ obtained from the first step, with the sparsity parameter $\tau_{s_j}$ and the fusion parameter $\tau_{f_j}$, and normalize the result to ensure a unit norm. 

In step 3, we update $\hat{w}^{(t+1)}_k$, given $\hat{\bbeta}^{(t+1)}_{k,1},\ldots,\hat{\bbeta}^{(t+1)}_{k,m+1}, \hat{\bbeta}^{(t)}_{k,m+2}$, which has a closed-form solution, 
\begin{equation} \label{eqn:what}
\hat{w}_k^{(t+1)} = \frac{\cR^{(t+1)} \times_1\hat{\bbeta}_{k,1}^{(t+1)} \times_2 \ldots \times_{m+1}\hat{\bbeta}_{k,m+1}^{(t+1)}}{\sum_{i=1}^{n} \left\{ \alpha_{i,k}^{(t)} \right\}^2 \left\| \Pi_{\Omega_i}\left( \hat{\bbeta}_{k,1}^{(t+1)} \circ \ldots \circ \hat{\bbeta}_{k,m+1}^{(t+1)} \right) \right\|_F^2},
\end{equation}
where $\cR^{(t+1)} = \sum_{i=1}^{n} \left\{ \alpha_{i,k}^{(t)} \right\}^2 \Pi_{\Omega_i}\left( \cR_{i,k}^{(t+1)} \right)$, and $\cR_{i,k}^{(t+1)}$ is as defined in \eqref{eqn:resid} by replacing $\hat{\bbeta}^{(t)}_{k,1},\ldots,\hat{\bbeta}^{(t)}_{k,m+1}$ with $\hat{\bbeta}^{(t+1)}_{k,1},\ldots,\hat{\bbeta}^{(t+1)}_{k,m+1}$. 

In step 4, we update $\hat{\bbeta}^{(t+1)}_{k,m+2}$, given $\hat{w}^{(t+1)}_k, \hat{\bbeta}^{(t+1)}_{k,1},\ldots,\hat{\bbeta}^{(t+1)}_{k,m+1}$, which again has a closed-form solution. Write $\tilde\cR_{i,k}^{(t+1)} = \cY_i - \sum_{{k'} \ne k, {k' }\in [r]} w_{k'}^{(t+1)} \alpha_{i,k'}^{(t)} \bbeta_{k',1}^{(t+1)} \circ \ldots \circ \bbeta_{k',m+1}^{(t+1)}$, and $\cA_k^{(t+1)} = w_k^{(t+1)} \bbeta_{k,1}^{(t+1)} \circ \ldots \circ \bbeta_{k,m+1}^{(t+1)}$. Then we have, 
\begin{equation} \label{eqn:betam2}
\hat{\bbeta}_{k,m+2}^{(t+1)} = \left\{ \frac{1}{n}\sum_{i=1}^{n} \left\| \Pi_{\Omega_i}\left( \cA_k^{(t+1)} \right) \right\|^2_F\xb_i\xb_i^{\top} \right\}^{-1}n^{-1}\sum_{i=1}^{n} \left\langle \Pi_{\Omega_i}\left( \tilde\cR_{i,k}^{(t+1)} \right), \Pi_{\Omega_i}\left( \cA_k^{(t+1)} \right) \right\rangle \xb_i,
\end{equation}
where $\langle \cdot, \cdot \rangle$ is the tensor inner product.

\change{We make some remarks regarding the convergence of Algorithm \ref{alg:STD}. First, with a suitable initial value, the iterative estimator from Algorithm \ref{alg:STD} is to converge to a neighborhood that is within the statistical precision of the true parameter at a geometric rate, as we show later in Theorems \ref{thm: rank1} and \ref{thm:generalr}. These results also provide a theoretical termination condition for Algorithm \ref{alg:STD}. That is, when the computational error is dominated by the statistical error, we can stop the algorithm. In practice, we iterate the algorithm until the estimates from two consecutive iterations are close, i.e., $\max_{j\in [m+2],k\in [r]} \min \left\{ \left\| \hat{\bbeta}_{k,j}^{(t+1)} - \hat{\bbeta}_{k,j}^{(t)} \right\|, \; \left\| \hat{\bbeta}_{k,j}^{(t+1)} + \hat{\bbeta}_{k,j}^{(t)} \right\| \right\} \leq 10^{-4}$. Second, with \emph{any} initial value, and if there are no sparsity and fusion constraints, i.e., without the \texttt{Truncatefuse} step, then Algorithm \ref{alg:STD} is guaranteed to converge to a stationary point, because the objective function monotonically decreases at each iteration \citep{WL20}. Finally, when imposing the sparsity and fusion constraints, the algorithmic convergence from any initial value becomes very challenging, since  both constraints are non-convex. Actually, the general convergence of non-convex optimizations remains an open question. For instance, in the existing non-convex models that employ truncation in optimizations, including sparse PCA \citep{ma2013sparse}, high-dimensional EM \citep{wang2015high}, sparse phase retrieval \citep{cai2016optimal}, sparse tensor decomposition \citep{sun2017provable}, and sparse generalized eigenvalue problem \citep{Tan2018sparse}, the convergence to a stationary point has only been established for a suitable initial value, but not for any initial value. We leave this as future research.}

The proposed Algorithm \ref{alg:STD} involves a number of tuning parameters, including the rank $r$, the sparsity parameter $\tau_{s_j}$, and the fusion parameter $\tau_{f_j}$, $j \in [m+1]$. We propose to tune the parameters by minimizing a BIC-type criterion,
\begin{eqnarray} \label{eqn: bic}
2\log\left\{ \frac{1}{n}\sum_{i = 1}^n \left\| \Pi_{\Omega_i} \left( \cY_i-\hat{\cB}\times_{m+2}\xb_i \right) \right\|_F^2 \right\} + \frac{\log\left( n\prod_{j=1}^{m+1}d_j \right)}{n\prod_{j=1}^{m+1}d_j}\times \textrm{df},
\end{eqnarray}
where the total degrees of freedom $\textrm{df}$ is the total number of unique nonzero entries of $\bbeta_{k,j}$. The criterion in $(\ref{eqn: bic})$ naturally balances the model fitting and model complexity.  Similar BIC-type criterions have been used in tensor data analysis  \citep{zhou2013, wang2015b, store2017}. To further speed up the computation, we tune the three sets of parameters $r$, $\tau_{s_j}$ and $\tau_{f_j}$ sequentially. That is, among the set of values for $r$, $\tau_{s_{j}}$, $\tau_{f_{j}}$, we first tune $r$ while fixing $\tau_{s_{j}},\, \tau_{f_{j}}$ at their maximum values. Then, given the selected $r$, we tune $\tau_{s_{j}}$, while fixing $\tau_{f_{j}}$ at its maximum value. Finally, given the selected $r$ and $\tau_{s_{j}}$, we tune $\tau_{f_{j}}$. In practice, we find such a sequential procedure yields good numerical performance.

\section{Theory}
\label{sec:theorem}

We next derive the non-asymptotic error bound of the actual estimator obtained from Algorithm \ref{alg:STD}. We first develop the theory for the case of rank $r = 1$, because this case has clearly captured the roles of various parameters, including the sample size, tensor dimension, and proportion of the observed entries, on both the computational and statistical errors. We then generalize to the case of rank $r>1$. We comment that, due to the involvement of the unobserved entries, our theoretical analysis is highly nontrivial, and is considerably different from \citet{store2017, SunLi2019}. We discuss in detail the effect of missing entries on both the regularity conditions and the theoretical properties. 

We first introduce the definition of the sub-Gaussian distribution. 
\change{\begin{definition}[sub-Gaussian]
The random variable $\xi$ is said to follow a sub-Gaussian distribution with a variance proxy $\sigma^2$, if $\mathbb{E}(\xi) = 0$, and for all $t \in \mathbb{R}$, $\mathbb{E}(\exp\{t\xi\})\le \exp\{t^2\sigma^2/2\}.$
\end{definition} }
Next we introduce some basic model assumptions common for both $r=1$ and $r>1$. Let $s_{j}$ denote the number of nonzero entries in $\bbeta^*_{k,j}$, $j\in[m+1]$, and $s = \max_j \{ s_j \}$.

\begin{assumption} \label{ass:model}
Assume the following conditions hold. 
\begin{enumerate}[(i)]
\item The predictor $\xb_i$ satisfies that $\|\xb_i\|\leq c_1$, $n^{-1}\sum_{i=1}^{n}\|\xb_i\xb^{\top}_i\|_2\le c_2$, $i\in [n]$, and  $1/c_0 < \lambda_{\min} \leq \lambda_{\max} < c_0$, where $\lambda_{\min}, \lambda_{\max}$ are the minimum and maximum eigenvalues of the sample covariance matrix $\Sigma = n^{-1}\sum_{i=1}^{n}\xb_i\xb^{\top}_i$, respectively, and $c_0, c_1, c_2$ are some positive constants. 

\item The true tensor coefficient $\cB^*$ in (\ref{eqn:true_model}) satisfies the CP decomposition \eqref{eqn:sparse_tensor} with sparsity and fusion constraints \eqref{eqn:parspace}, and the decomposition is unique up to a permutation. Moreover, $\|\cB^*\|\le c_3w^*_{\max}$ where $w^*_{\max} = \max_k\{w^*_k\}$, $w^*_{\min} = \min_k\{w^*_k\}$, and $c_3$ is some positive constant. Furthermore, $w^*_{\max} = \mathcal{O}(w^*_{\min})$.

\item The decomposed component $\bbeta^*_{k,j}$ is a $\mu$-mass unit vector, in that $\max_{l\in d_j}|\bbeta^*_{k,j,l}|\le \mu/\sqrt{s}$.

\item \change{The entries in the error tensor $\cE_i$ are i.i.d.\ sub-Gaussian with a variance proxy $\sigma^2$. }

\item The entries of the dynamic tensor response $\cY_i$ are observed independently with an equal probability $p \in (0,1]$. 
\end{enumerate}
\end{assumption}

\noindent
We make some remarks about these conditions. Assumption \ref{ass:model}(i) is placed on the design matrix, which is mild and can be easily verified when $\xb_i$ is of a fixed dimension. Assumption \ref{ass:model}(ii) is about the key structures we impose on the coefficient tensor $\cB^*$. It also ensures the identifiability of the decomposition of $\cB^*$, which is always imposed in CP decomposition based tensor analysis \citep{zhou2013, store2017,ChenYuan2019}.  Assumption \ref{ass:model}(iii) is to ensure that the mass of the tensor would not concentrate on only a few entries. In that extreme case, randomly observed entries of the tensor response may not contain enough information to recover $\cB^*$. Note that, since $\bbeta^*_{k,j}$ is a vector of unit length, a relatively small $\mu$ implies that the nonzero entries in $\bbeta^*_{k,j}$ would be more uniformly distributed. This condition has been commonly imposed in the tensor completion literature for the same purpose \citep{JO2014}. \change{Assumption \ref{ass:model}(iv) assumes the error terms \change{follow a sub-Gaussian distribution}. This assumption is again fairly common in theoretical analysis of tensor models \citep{clpc19, xyz20}.} Finally, Assumption \ref{ass:model}(v) specifies the mechanism of how each entry of the tensor response is observed, which is assumed to be independent of each other and have an equal observation probability. We recognize that this is a relatively simple mechanism. It may not always hold in real applications, as the actual observation pattern of the tensor data can depend on multiple factors, and may not be independent for different entries. We impose this condition for our theoretical analysis, even though our estimation algorithm does not require it. In the tensor completion literature, this mechanism has been commonly assumed \citep{JO2014, yuan2016on, yuan2017incoherent, dong2017}. We have chosen to impose this assumption because the theory of supervised tensor learning even for this simple mechanism remains unclear, and is far from trivial. We feel a rigorous theoretical analysis for this mechanism itself deserves a full investigation. We leave the study under a more general observation mechanism as future research.

\subsection{Theory with $r=1$}
\label{sec:theoryr1}

To ease the notation and simplify the presentation, we focus primarily on the case with a third-order tensor response, i.e., $m = 2$. This however does not lose generality, as all our results can be extended to the case of $m>2$ in a straightforward fashion. Let $d = \max\{d_1,\cdots,d_{m+1}\}$. Next, we introduce some additional regularity conditions. 

\begin{assumption} \label{ass:missing}
Assume the observation probability $p$ satisfies that, 
\change{\begin{eqnarray*}
p \geq \frac{c_4\{\log (d)\}^4 \mu^3}{n\,s^{1.5}}.
\end{eqnarray*}}
where $c_4 > 0$ is some constant.
\end{assumption}

\noindent
Due to Assumption \ref{ass:model}(v), the observation probability $p$ also reflects the proportion of the observed entries of the tensor response. Assumption \ref{ass:missing} places a lower bound on this proportion to ensure a good recovery of the tensor coefficient. This bound depends on the sample size $n$, true sparsity parameter $s$, maximum dimension $d$, and mass parameter $\mu$. \change{We discuss these dependencies in detail. First, compared to the lower bound conditions on $p$ in the tensor completion literature  where a single tensor is considered \citep{JO2014, yuan2016on, yuan2017incoherent,dong2017, clpc19}, our lower bound is different, as it depends on the number of tensor samples $n$,  and it tends to $0$ as $n$ tends to infinity. When $n=1$, our lower bound is comparable to that in \cite{JO2014, clpc19}, with $s$ replaced by $d$, as they did not consider any sparsity. Second, the lower bound on $p$ increases as $s$ decreases, i.e., as the data becomes more sparse. This is because, when the sparsity is involved, both our problem and the tensor completion problem become more difficult. Intuitively, when the sparsity increases, the nonzero elements may concentrate on only a few tensor entries. As a result, a larger proportion of the tensor entries needs to be observed to ensure that a sufficient number of nonzero elements can be observed for tensor estimation or completion. We also note that this condition on the lower bound on $p$ is different from the sample complexity condition on $n$ that we will introduce in Assumption \ref{ass:samplesize}. The latter suggests that the required sample size $n$ decreases as $s$ decreases. Third, when there is no sparsity, \cite{JO2014, clpc19} showed that the lower bound on $p$ is of the order $(\log d)^4 / (d^{3/2})$, which decreases as $d$ increases. In our setting with the sparsity, however, the lower bound on $p$ increases as $d$ increases. Finally, the lower bound on $p$ increases as the mass parameter $\mu$ increases. This is because when $\mu$ increases, the mass of the tensor may become more likely to concentrate on a few entries, and thus the entries need to be observed with a larger probability to ensure the estimation accuracy.}

\begin{assumption} \label{ass:minigap}
Assume the sparsity and fusion parameters satisfy that $\tau_{s_j} \ge s_{j}, \tau_{s_j} =\mathcal{O}(s_j)$, and $\tau_{f_j} \ge f_{j}$. Moreover, define the minimal gap, $\Delta^*=\min_{1< s\le d_{j},  \bbeta^*_{1,j,s}\ne \bbeta^*_{1,j,s-1}, j\in [3]}|\bbeta^*_{1,j,s}-\bbeta^*_{1,j,s-1}|$. Assume that, for the positive constant $C_1$ as defined in Theorem \ref{thm: rank1}, we have
\begin{eqnarray*}
\change{\Delta^* > \frac{C_1 \sigma}{w^*_1}\sqrt{\frac{s\log(d)}{np}}.}
\end{eqnarray*}
\end{assumption}

\noindent
The condition for the sparsity parameter ensures that the truly nonzero elements would not be shrunk to zero. Similar conditions have been imposed in truncated sparse models \citep{yuan2013, wang2015high, sun2017provable, Tan2018sparse}. The conditions for the fusion parameter and the minimum gap ensure that the fused estimator would not incorrectly merge two distinct groups of entries in the true parameter. Such conditions are common in sparse and fused regression models \citep{Tibshirani2005, rinaldo2009}.

\begin{assumption}\label{ass:initialization}
Define the initialization error $\epsilon = \max\{|\widehat{w}^{(0)}_1 - w_1^*|/w^*_1 ,\, \max_{j} \|\widehat{\bbeta}_{1,j}^{(0)}- \bbeta_{1,j}^*\|_2 \} $.  Assume that
\vspace{-0.15in}
\begin{eqnarray*}
\epsilon <  \min\left\{\frac{\lambda_{\min}^3}{24\sqrt{10}\,c_2\, \lambda_{\max}^2}, \; \frac{1}{6}\right\},
\end{eqnarray*}
where $c_2$ is the same constant as in Assumption \ref{ass:model}. 
\end{assumption}

\noindent
This assumption is placed on the initialization error of Algorithm \ref{alg:STD}, and requires that the initial values are reasonably close to the true parameters. Particularly, the condition on $\epsilon$ requires the initial error to be smaller than some constant, which is a relatively mild condition, since $\bbeta_{1,j}^*$'s are unit vectors. Such constant initialization condition is commonly employed in the tensor literature \citep{store2017, hwz20, xyz20}. In Section \ref{sec:initialization}, we further propose an initialization procedure, and show both theoretically and empirically that such a procedure can produce initial values that satisfy Assumption \ref{ass:initialization}. 

\begin{assumption}\label{ass:samplesize}
\change{Assume the sample size $n$ satisfies that 
\begin{eqnarray*}
n \ge \max\left\{ \frac{c_5\,\sigma^2\,s^2\,\log(d)}{w^{*2}_1p}, \, \frac{c_6\sigma s\log(d)\log(\sqrt{s^3/p})}{w^*_1p} \right\}
\end{eqnarray*}
where $c_5$ and $c_6$ are some positive constants.}
\end{assumption}

\noindent
\change{There are two terms in this lower bound, both of which are due to the error tensor $\cE_i$ in the model and the missing entries in the response tensor. In addition, the first term is needed to ensure the $\mu$-mass condition is satisfied. When the observational probability $p$ satisfies the lower bound requirement in Assumption \ref{ass:missing}, the required sample size decreases as $s$ decreases, since in this case the number of free parameters decreases. When the strength of signal $w^*_1$ increases or the noise level $\sigma$ decreases, the required sample size also decreases.}

We now state the main theory for the estimator of Algorithm \ref{alg:STD} when $r=1$. 

\begin{theorem}
\label{thm: rank1}
Suppose Assumptions \ref{ass:model}-\ref{ass:samplesize} hold. When the tensor rank $r=1$, the estimator from the $t$-th iteration of Algorithm \ref{alg:STD} satisfies that, with high probability,	
\change{\begin{eqnarray*}
& &  \max \left\{ |\widehat{w}^{(t)}_1 - w_1^*|/w^*_1,  \;  \max_{j} \|\widehat{\bbeta}_{1,j}^{(t)} -  \bbeta_{1,j}^*\|_2 \right\} \\
& \le & \underbrace{ \kappa^t \epsilon}_{\textrm{computational error}} \; + \;\;\; \underbrace{\frac{1}{1-\kappa} \frac{C_1\sigma}{w^*_1}\sqrt{\frac{s\log(d)}{np}}}_{\textrm{statistical error}},
\end{eqnarray*}}
where $\kappa =6\sqrt{10}c_2\lambda_{\max}^2 \epsilon / \lambda_{\min}^3+1/2< 1$ is the positive contraction coefficient, with $\epsilon$ as defined in Assumption \ref{ass:initialization}, and the constant $C_1= (6\sqrt{10}\tilde{C}\lambda_{\max}+\tilde{C}_2\lambda_{\min}\sqrt{q}) / \lambda^2_{\min}$. Here $c_2$ is the same constant as defined in Assumptions \ref{ass:model},  $\tilde{C},\tilde{C}_2$ are some positive constants, and $q$ is fixed under Assumption \ref{ass:model}$(i)$.
\end{theorem}

\noindent	
The non-asymptotic error bound in Theorem \ref{thm: rank1} can be decomposed as the sum of a computational error and a statistical error. The former is related to the optimization procedure, while the latter is related to the statistical model. The statistical error decreases with a decreasing $\kappa$, an increasing signal to noise ratio as reflected by $\sigma/w^*_1$, an increasing sample size $n$ and an increasing observation probability $p$. \change{When $p=1$ and $\sigma=1$, the statistical error rate in our Theorem \ref{thm: rank1} actually improves the statistical error rate in the completely observed tensor response regression \citep{store2017}, which is of order $1/w^*_1\sqrt{s^3\log(d)/n}$. This improvement is achieved because we have employed a new proof technique using the covering number argument \citep{ts14} in bounding the sparse spectral of the error tensor, which allows us to obtain a sharper rate in terms of the sparsity parameter $s$. Moreover, when $n=1$ and $s=d$, our statistical error rate matches with the rate $\sigma/w^*_1\sqrt{d\log(d)/p}$ in the non-sparse tensor completion \citep{clpc19}. }

One of the key challenges of our theoretical analysis is the complicated form of the element-wise estimator $\tilde{\bbeta}_{k,3}$ in \eqref{eqn:betanew}. Consequently, one cannot directly characterize the distance between $\tilde{\bbeta}_{k,3}/\|\tilde{\bbeta}_{k,3}\|$ and ${\bbeta}^*_{k,3}$ with a simple analytical form. Furthermore, the presence of noise error poses several fundamental challenges. The missing entries in noise tensors make existing proof techniques no longer applicable in our theoretical analysis. As we shall demonstrate later, we need to carefully control the upper bound of error tensor with missing entries. 

We also briefly comment that, Theorem \ref{thm: rank1} provides a theoretical termination condition for Algorithm \ref{alg:STD}. When the number of iterations $t$ exceeds $O\{\log_{1/\kappa}(\epsilon / \varepsilon^*) \}$, where $\varepsilon^*$ is the statistical error term in Theorem \ref{thm: rank1}, then the computational error is to be dominated by the statistical error, and the estimator falls within the statistical precision of the true parameter.

\subsection{Theory with $r > 1$}
\label{sec:theoryrgeneral}

Next, we extend our theory to the general rank $r > 1$. The regularity conditions for the general rank case parallel those for the rank one case. Meanwhile, some modifications are needed, due to the interplay among different decomposed components $\bbeta_{k,j}$.

\begin{assumption} \label{ass:missing_r}
Assume the observation probability $p$ satisfies that 
\change{\begin{eqnarray*}
p\geq \frac{c_7\{\log(d)\}^4 \mu^3rw^{*2}_{\max}}{n\,s^{1.5}\,w^{*2}_{\min}},
\end{eqnarray*}}
where  $c_7 > 0$ is some constants.
\end{assumption}

\noindent
For the general rank case, the lower bound on the observation probability $p$ depends additionally on the rank $r$ and the ratio $w^*_{\max}/w^*_{\min}$. In particular, it is to increase with an increasing rank $r$, which suggests that more observations are needed if the rank of the coefficient tensor increases. When the sample size $n=1$, our condition is comparable to that in tensor completion \citep{JO2014}, where the latter requires $p\ge c\mu^6r^5w^{*4}_{\max} / (d^{1.5}w^{*4}_{\min})$ ignoring the logarithm term, with $s$ replaced by $d$, as they do not consider any sparsity.

\begin{assumption} \label{ass:minigap_r}
Assume the sparsity and fusion parameters satisfy that $\tau_{s_j} \ge s_{j}, \tau_{s_j} =\mathcal{O}(s_j)$, and $\tau_{f_j} \ge f_{j}$. Moreover, define the minimal gap $\Delta^*=\min_{1< s\le d_{j},  \bbeta^*_{k,j,s}\ne \bbeta^*_{k,j,s-1}, j\in [3], k\in[r], } |\bbeta^*_{k,j,s}-\bbeta^*_{k,j,s-1}|$. Assume that, 
\begin{eqnarray*}
\change{\Delta^*> \frac{C_2\sigma w^*_{\max}}{w^{*2}_{\min}}\sqrt{\frac{s\log(d)}{np}},}
\end{eqnarray*}
where positive constant $C_2$ is the same constant as defined in Theorem \ref{thm:generalr}.
\end{assumption}

\noindent
This assumption is similar to Assumption \ref{ass:minigap}, and it reduces to Assumption \ref{ass:minigap} when $r=1$. 

\begin{assumption} \label{ass:initialization_r}
Define $\epsilon = \max_k\left\{|\hat{w}^{(0)}_k-w^*_k|/ w^*_{k},\, \max_{j} \|\widehat{\bbeta}_{k,j}^{(0)} -  \bbeta_{k,j}^*\|_2\right\}$. Assume $\epsilon$ satisfies,  
\begin{eqnarray*}
\epsilon < \min\left\{ \frac{\lambda_{\min}^3w^{*2}_{\min}}{24\sqrt{10}c_2\lambda^2_{\max}w^{*2}_{\max}r},   \; \frac{\lambda_{\min}^3w^{*3}_{\min}}{4c_1^2c_2\lambda_{\max}w^{*3}_{\max}r^2},\; \frac{1}{6} \right\},
\end{eqnarray*}
where $c_1, c_2$ are the same constants as defined in Assumption \ref{ass:model}.
\end{assumption}

\noindent
It is seen that the initial error depends on the rank $r$.  The upper bound tightens as $r$ increases, as in such a case, the tensor recovery problem becomes more challenging. It is also noted that, when $r = 1$, this condition is still stronger than that in Assumption \ref{ass:initialization}. This is due to the interplay among different decomposed components in general rank case. 

\begin{assumption}
\label{ass:incoherence} 
Define the incoherence parameter $\xi = \max_{j\in[3], k\ne k^{\prime}} \left| \langle \bbeta^*_{k,j},\bbeta^*_{k^{\prime},j} \rangle \right|$. Assume,
\begin{eqnarray*}
\xi\le \frac{\lambda_{\min}^3w_{\min}^{*3}}{4c_1^2c_2\lambda_{\max}w_{\max}^{*3}r^{2}},
\end{eqnarray*}
where $c_1, c_2$ are the same constants as defined in Assumption \ref{ass:model}. 
\end{assumption}

\noindent
For the general rank case, we need to control the correlations between the decomposed components across different ranks. The incoherence parameter $\xi$ quantifies such correlations. As rank $r$ increases, the upper bound on $\xi$ becomes tighter. Similar conditions have been introduced in \cite{anandkumar2014a, sun2017provable, hzc20}.

\begin{assumption} \label{ass:samplesize_r}
\change{Assume the sample size $n$ satisfies that,
\begin{eqnarray*}
n \ge \max\left\{\frac{c_5\,\sigma^2\,s^2\,\log(d)}{w^{*2}_{\min}p}, \, \frac{c_6\sigma s\log(d)\log(\sqrt{s^3/p})}{w^*_{\min}p} \right\}
\end{eqnarray*}
where $c_5$ and $c_6$  are the same positive constants as defined in Assumption \ref{ass:samplesize}. }
\end{assumption}
\noindent
This assumption is similar to Assumption \ref{ass:samplesize}, and it reduces to Assumption \ref{ass:samplesize} when $r=1$.  

We next state the main theory for the estimator of Algorithm \ref{alg:STD} when $r > 1$. 

\begin{theorem}
\label{thm:generalr}
Suppose Assumptions \ref{ass:model} and \ref{ass:missing_r}-\ref{ass:samplesize_r} 
hold. For a general rank $r$, the estimator from the $t$-th iteration of Algorithm \ref{alg:STD} satisfies that, with a high probability, 
\change{\begin{eqnarray*}
& & \max \left\{\max_k|\widehat{w}^{(t)}_k - w_k^*|/w^*_k, \; \max_{k,j} \|\widehat{\bbeta}_{k,j}^{(t)} -  \bbeta_{k,j}^*\|_2 \right\} \\
& \le & \underbrace{\tilde{\kappa}^t \epsilon}_{\text{computational error}} + \underbrace{\frac{1}{1-\tilde{\kappa}} \frac{C_2w^{*}_{\max}\sigma}{w^{*2}_{\min}}\sqrt{\frac{s\log(d)}{np}}}_{\text{statistical error}}.
\end{eqnarray*}}
where 
\begin{eqnarray*}
\tilde{\kappa} = 
\frac{6\sqrt{10}c_2\lambda_{\max}^2w^{*2}_{\max}r}{\lambda_{\min}^3w^{*2}_{\min}}\epsilon+\frac{c_1^2c_2\lambda_{\max}w^{*3}_{\max}r^2}{\lambda_{\min}^3w^{*3}_{\min}}\epsilon+\frac{c_1^2c_2\lambda_{\max}w^{*3}_{\max}r^2}{\lambda_{\min}^3w^{*3}_{\min}}\xi+\frac{1}{4}< 1, 
\end{eqnarray*}
is the positive contraction coefficient, and the constants $C_2 = (6\sqrt{10}\tilde{C}\lambda_{\max}+12\tilde{C}_2\sqrt{q}\lambda_{\min})/\lambda^2_{\min}$. Here $c_1, c_2$ is the same constant as defined in Assumptions \ref{ass:model}, $\tilde{C},\tilde{C}_2$ are some positive constants, and $q$ is fixed under Assumption \ref{ass:model}$(i)$.

\end{theorem}

\noindent
The contraction coefficient $\tilde{\kappa}$ is greater than $\kappa$ in Theorem \ref{thm: rank1}, which indicates that the algorithm has a slower convergence rate for the general rank case. Moreover, $\tilde{\kappa}$ increases with an increasing rank $r$. This agrees with the expectation that, as the tensor recovery problem becomes more challenging, the algorithm is to have a slower convergence rate.

\subsection{Initialization}
\label{sec:initialization}

As the optimization problem in (\ref{eqn:objective}) is nonconvex, the success of Algorithm \ref{alg:STD} replies on good initializations. Motivated by \cite{clpc19}, we next propose a spectral initialization procedure for $r=1$ and $r>1$, respectively. Theoretically, we show that, the produced initial estimator satisfies the initialization Assumption \ref{ass:initialization} when $r=1$. Numerically, we demonstrate that the initialization error decays fast for both $r=1$ and $r>1$ cases as the sample size $n$ increases, and thus the constant initialization error bound in the initialization Assumptions \ref{ass:initialization} and \ref{ass:initialization_r} is expected to hold with a sufficiently large $n$.

\begin{algorithm}[b!]
\caption{Spectral initialization algorithm for $r=1$.}
\begin{algorithmic}[1]
\STATE \textbf{input:} the number of restarts $L$,  the estimates $\Ub_1,\, \Ub_2$, and the sparsity parameter $\tau_{s_{j}},\,j\in[3]$.
\FOR{$l=1$ to $L$}
\STATE generate $\gb_1^{l}\sim \textrm{Normal}(\mathbf{0},\mathbf{I}_{d_3})$, and compute $\tilde{\gb}^{l}_1=\Ub_1\Ub_1^{\top}\gb_1$, $\Mb_1^{l} =p^{-1}\cT \times_3 \tilde{\gb}_1^{l}$. \label{line:gl}
\STATE set $\vb^{l}_1$ and $\vb^{l}_2$ as the first left and right singular vector of $\Mb_1^{l}$ corresponding to the largest absolute singular value $|\lambda^{l}_1|$. 
\ENDFOR
\FOR{$l=1$ to $L$}
\STATE generate $\gb_2^{l}\sim \textrm{Normal}(\mathbf{0},\mathbf{I}_{d_1})$, and compute $\tilde{\gb}^{l}_2=\Ub_2\Ub_2^{\top}\gb_2$, $\Mb_2^{l} =p^{-1}\cT \times_3 \tilde{\gb}_2^{l}$. 
\STATE set $\vb^{l}_3$ and $\vb^{l}_4$ as the left and right singular vector of $\Mb_2^{l}$ corresponding to the largest absolute singular value $|\lambda^{l}_2|$. 
\ENDFOR
\STATE choose $(\vb_1,\vb_2)$ from $\{(\vb_1^l,\vb_2^l)\}_{l=1}^{L}$ with the largest $|\lambda^{l}_1|$; choose $(\vb_3,\vb_4)$ similarly. 
\STATE compute $\widehat{\bbeta}^{(0)}_{1,j} = \texttt{Norm}(\texttt{Truncate}(\tilde{\vb}_j,\tau_{s_j}))$ for $j = 1, 2, 3$, where $(\tilde{\vb}_1,\,\tilde{\vb}_2,\,\tilde{\vb}_3)$ is obtained from $(\vb_1,\vb_2)$, $(\vb_3,\vb_4)$, and $\texttt{Norm}$ is the normalization operator. \label{line:beta123}
\STATE compute $\hat{w}_1^{(0)}$ and $\hat{\bbeta}_{1,4}^{(0)}$ using (\ref{eqn:betaw}).
\STATE\textbf{output:} $\hat{w}_1^{(0)},\, \widehat{\bbeta}^{(0)}_{1,1},\, \widehat{\bbeta}^{(0)}_{1,2},\, \widehat{\bbeta}_{1,3}^{(0)}$ and $\widehat{\bbeta}_{1,4}^{(0)}$.
\end{algorithmic}
\label{alg:initialization2}
\end{algorithm}

We first present the initialization procedure for $r=1$ in Algorithm \ref{alg:initialization2}. Denote $\cT = n^{-1}\sum_i\Pi_{\Omega_{i}}(\cY_i)$. Let $\Ab_1=\text{unfold}_3(p^{-1}\cT) \in \mathbb{R}^{d_3\times d_1d_2}$, and $\Bb_1 = \Pi_{\text{off-diag}}(\Ab_1\Ab_1^{\top})\in\mathbb{R}^{d_3\times d_3}$, where $\Pi_{\text{off-diag}}(\cdot)$ keeps only the off-diagonal entries of the matrix. Let $\Ub_1 \bLambda_1 \Ub_1^{\top}$ be the rank-$r$ decomposition of $\Bb_1$. Next, let $\Ab_2 = \text{unfold}_1(p^{-1}\cT) \in \mathbb{R}^{d_1\times d_2d_3}$, $\Bb_2 =\Pi_{\text{off-diag}}(\Ab_2 \Ab_2^{\top})\in\mathbb{R}^{d_1\times d_1}$, and $\Ub_2 \bLambda_2 \Ub_2^{\top}$ be the rank-$r$ decomposition of $\Bb_2$. We then feed $\Ub_1$ and $\Ub_2$ into Algorithm \ref{alg:initialization2}. When $r=1$, we have $E(\Ab_1) = w_1^*\sum_i\frac{1}{n}(\bbeta_{1,4}^{*\top}\xb_i)\bbeta_{1,3}^*(\bbeta^*_{1,1}\otimes \bbeta^*_{1,2})^{\top}$, whose column space is the span of $\bbeta_{1,3}^*$. A natural way to estimate the column space of $E(\Ab_1)$ is from the principal space of $\Ab_1 \Ab_1^{\top}$. Similar to \cite{clpc19}, we exclude the diagonal entries of $\Ab_1\Ab_1^{\top}$ to remove their influence on the principal directions. To retrieve tensor factors from the subspace estimate, we first generate random vectors from normal distribution, i.e., $\gb_1^l$ in line \ref{line:gl} and $\gb_2^l$ in line 7 of Algorithm \ref{alg:initialization2} . Then we project the random vectors $\gb^l_1$ and $\gb^l_2$ onto $\Ub_1$ and $\Ub_2$. This projection step helps mitigate the perturbation incurred by both unobserved values and data noise \citep{clpc19}. Note that $E(\Mb^l_1\mid\tilde{\gb}^l_1) = w^*_1\sum_{i}n^{-1} (\bbeta_{1,4}^{*\top}\xb_{i})\langle\bbeta_{1,3}^{*}, \tilde{\gb}^l_1\rangle\bbeta_{1,1}^*\bbeta_{1,2}^{*\top}$. Correspondingly, the left leading singular vector corresponds to the largest absolute singular value of $\Mb^l_1$ is expected to be close to $\bbeta_{1,1}^*$. Similarly, the right leading singular vector of $\Mb^l_1$ is expected to be close to $\bbeta_{1,2}^*$. Following the same argument, we can obtain a good estimate of $\bbeta_{1,2}^*$ and $\bbeta_{1,3}^*$ from $\Mb^l_2$.  Then, in line \ref{line:beta123} of Algorithm \ref{alg:initialization2}, we match the identified singular vector pairs with $(\tilde{\vb}_1,\tilde{\vb}_2,\tilde{\vb}_3)$. That is, let $l=\arg\max_{j=3,4}\left\{\max_{k=1,2}\{\langle\vb_j,\vb_k\rangle\}\right\}$. Set $\tilde{\vb}_2=\vb_l$, the remaning one in the pair $(\vb_3,\vb_4)$ as $\tilde{\vb}_3$, and $\tilde{\vb}_1=\argmin_{j = 1,2} \{\langle\vb_j,\tilde{\vb}_2\rangle\}$. Next, given $\hat{\bbeta}^{(0)}_{1,1} ,\, \hat{\bbeta}^{(0)}_{1,2},\, \hat{\bbeta}^{(0)}_{1,3}$, we obtain $\hat{w}^{(0)}_1,\, \hat{\bbeta}_{1,4}^{(0)}$ by solving the following optimization,
\begin{equation*} \label{eqn:obj_add}
\min\nolimits_{\substack{w_1>0, \|\bbeta_{1,4}\| =1}} \; 
\frac{1}{n}\sum_{i = 1}^n \left\| \Pi_{\Omega_i}\left( \cY_i -  w_1 (\bbeta_{1,4}^{\top} \xb_i) \hat{\bbeta}^{(0)}_{1,1} \circ \hat{\bbeta}^{(0)}_{1,2} \circ \hat{\bbeta}^{(0)}_{1,3} \right)  \right\|_F^2.
\end{equation*}
Finally, leting $\cA = \hat{\bbeta}_{1,1}^{(0)}\circ \hat{\bbeta}_{1,2}^{(0)}\circ\hat{\bbeta}_{1,3}^{(0)}$, we obtain the initial estimates $\hat{\bbeta}_{1,4}^{(0)}$ and $\hat{w}^{(0)}_{1}$ as
\begin{align} \label{eqn:betaw}
\begin{split}
\hat{\bbeta}^{(0)}_{1,4} & = \text{Norm}\left(\left\{\frac{1}{n}\sum_i\left\| \Pi_{\Omega_{i}}(\cA)\right\|_F^2\xb_{i}\xb_{i}^{\top}\right\}^{-1}n^{-1}\sum_{i}\langle\Pi_{\Omega_{i}}(\cY_i), \Pi_{\Omega_i}(\cA)\rangle\xb_{i}\right), \\
\hat{w}^{(0)}_{1} & = \frac{\sum_i\hat{\bbeta}_{1,4}^{(0)\top}\xb_{i}\Pi_{\Omega_{i}}(\cY_i)\times_1\hat{\bbeta}^{(0)}_{1,1}\times_2\hat{\bbeta}^{(0)}_{1,2}\times_3\hat{\bbeta}^{(0)}_{1,3}}{\sum_i\{\hat{\bbeta}_{1,4}^{(0)\top}\xb_i\}^2\left\|\Pi_{\Omega_{i}}\left(\cA\right)\right\|_F^2}. 
\end{split}
\end{align}

We next present the initialization procedure for $r>1$ in Algorithm \ref{alg:initialization3}. We first apply Algorithm \ref{alg:initialization2} to generate two sets ${(\vb_1^l,\vb_2^l)}_{l=1}^L$, ${(\vb_3^l,\vb_4^l)}_{l=1}^L$. Since $\hat{\bbeta}_{k,1}$ and $\hat{\bbeta}_{k,2}$ are from $(\vb_1^l,\vb_2^l)$, and $\hat{\bbeta}_{k,2}, \hat{\bbeta}_{k,3}$ are from $(\vb_3^l,\vb_4^l)$, we merge the two and find the triplet $(\tilde{\vb}_1^l, \tilde{\vb}_2^l, \tilde{\vb}_3^l)$. Next, we search for $(\hat{\bbeta}_{k,1},\hat{\bbeta}_{k,2},\hat{\bbeta}_{k,3})$ such that $| p^{-1} \cT \times_1\tilde{\vb}_1 \times_2\tilde{\vb}_2\times_3\tilde{\vb}_3| $ is maximized. This is because the selected vectors are expected to be close to true factors when $| p^{-1} \cT \times_1\tilde{\vb}_1 \times_2\tilde{\vb}_2\times_3\tilde{\vb}_3| $ is large \citep{sun2017provable}. We also remove all those triplets that are close to $(\hat{\bbeta}_{k,1},\hat{\bbeta}_{k,2},\hat{\bbeta}_{k,3})$, since they eventually generate the same decomposition vectors up to the tolerance parameter. We then iteratively refine the selected vectors. In our numerical experiments, we have found that one iteration is often enough, while the algorithm is not sensitive to the tolerance parameter $\epsilon_{th}$ neither due to the refinement step.

\begin{algorithm}[t!]
\caption{Spectral initialization algorithm for $r>1$.}
\begin{algorithmic}[1]
\STATE \textbf{input:} the number of restarts L, the estimates $\Ub_1,\, \Ub_2$, the tolerance parameter $\epsilon_{th}$, and the sparsity parameter $\tau_{s_{j}},\,j\in[3]$.
\STATE obtain ${(\vb_1^l,\vb_2^l)}_{l=1}^L$, ${(\vb_3^{l},\vb_4^l)}_{l=1}^L$ using Algorithm \ref{alg:initialization2}. 
\STATE obtain the triplet $S = \{(\tilde{\vb}_1^l, \tilde{\vb}_2^l, \tilde{\vb}_3^l)\}_{l=1}^L$ from ${(\vb_1^l,\vb_2^l)}_{l=1}^L$, ${(\vb_3^{l},\vb_4^l)}_{l=1}^L$.
\FOR{$k=1$ to $r$}
\STATE find $(\hat{\bbeta}_{k,1},\hat{\bbeta}_{k,2}, \hat{\bbeta}_{k,3}) = \arg\max_{(\tilde{\vb}_1^l,\tilde{\vb}_2^l,\tilde{\vb}_3^l)\in S} |p^{-1}T\times_1\tilde{\vb}_1^l\times_2\tilde{\vb}_2^l\times_3\tilde{\vb}_3^l|$. 
\STATE remove all triplets in ${(\tilde{\vb}_1^l, \tilde{\vb}_2^l, \tilde{\vb}_3^l)}_{l=1}^L$ with $\max\{|\langle\hat{\bbeta}_{k,1}, \tilde{\vb}_1^l\rangle|, |\langle\hat{\bbeta}_{k,2},\tilde{\vb}_2^l\rangle|, |\langle\hat{\bbeta}_{k,3},\tilde{\vb}_3^l\rangle|\}>1-\epsilon_{th}$. 
\ENDFOR
\STATE set $\hat{w}_k=1$, and randomly generate unit-norm vectors $\hat{\bbeta}_{k,4},\, k\in[r]$ from a standard normal distribution. 
\REPEAT
\STATE update $ \hat{\bbeta}_{k,1}, \hat{\bbeta}_{k,2},\hat{\bbeta}_{k,3}$ using \eqref{eqn:betanew}, and set $\hat{\bbeta}_{k,j} = \texttt{Norm}(\texttt{Truncate}(\hat{\bbeta}_{k,j}.\tau_{s_{j}})), j\in[3]$, 
\STATE update $\hat{w}_k$ using \eqref{eqn:what}, and update $\hat{\bbeta}_{k,4}$ using \eqref{eqn:betam2}, $k\in[r]$. 
\UNTIL the stopping criterion is met 
\STATE denote the final update of $\hat{w}_k, \{\hat{\bbeta}_{k,j}\}_{j=1}^4$ as $\hat{w}_k^{(0)}, \{\hat{\bbeta}_{k,j}^{(0)}\}_{j=1}^4$, $k\in [r]$, respectively. 
\STATE\textbf{output:} $\hat{w}_k^{(0)},\, \widehat{\bbeta}_{k,1}^{(0)},\, \widehat{\bbeta}_{k,2}^{(0)},\, \widehat{\bbeta}_{k,3}^{(0)}, \, \widehat{\bbeta}_{k,4}^{(0)},\, k\in[r]$. 
\end{algorithmic}
\label{alg:initialization3}
\end{algorithm}

Next, we present a proposition showing that the initial estimator obtained from Algorithm \ref{alg:initialization2} satisfies the initialization Assumption \ref{ass:initialization} when $r=1$. The theoretical guarantee for the $r>1$ case is very challenging, and we leave it as future research. 

\begin{proposition} \label{thm:initial}
Suppose Assumptions \ref{ass:model}, \ref{ass:missing}, \ref{ass:minigap}, and \ref{ass:samplesize} hold. Furthermore, suppose $L\ge C^{\prime}_1$ for some large enough $C^{\prime}_1$, $|\sum_{i}n^{-1}\bbeta_{1,4}^{*\top}\xb_{i}|\ge C^{\prime}_2$ for some constant $C^{\prime}_2>0$. Then, the initial estimator produced by Algorithm \ref{alg:initialization2} satisfies that 
\begin{eqnarray*}
\change{\max\left\{|\widehat{w}^{(0)}_1 - w_1^*|/w^*_1, \max_{j} \|\widehat{\bbeta}_{1,j}^{(0)} -  \bbeta_{1,j}^*\|_2 \right\}
= O_p\left\{ \sqrt{\frac{\log(d)}{nps^2}}+\frac{\sigma}{w^*_1}\sqrt{\frac{s\log(d)}{np}}\right\}.}
\end{eqnarray*}
\end{proposition}
\noindent
We make a few remarks. First, this result shows that the error of the initial estimator obtained from Algorithm \ref{alg:initialization2} decays with $n$, and thus the constant initialization error bound on $\epsilon$ in Assumption \ref{ass:initialization} is guaranteed to hold as $n$ increases. Second, the estimation error in Proposition \ref{thm:initial} is slower than the statistical error rate in Theorem \ref{thm: rank1} when $\sigma/w^*_1\le c/s^{1.5}$. This suggests that, after obtaining the initial estimator from Algorithm \ref{alg:initialization2}, applying the alternating block updating Algorithm \ref{alg:STD} could further improve the error rate of the estimator. 

\begin{figure}[t!]
\caption{Estimation error of the initial estimator by the spectral initialization algorithms as the sample size increases. The left panel is for $r=1$, and the right panel is for $r=2$.}
\vspace{-0.15in}
\centering
\begin{tabular}{cc}
\includegraphics[scale = 0.45]{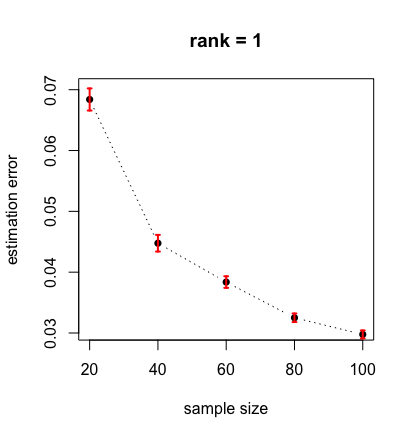} & 
\includegraphics[scale = 0.45]{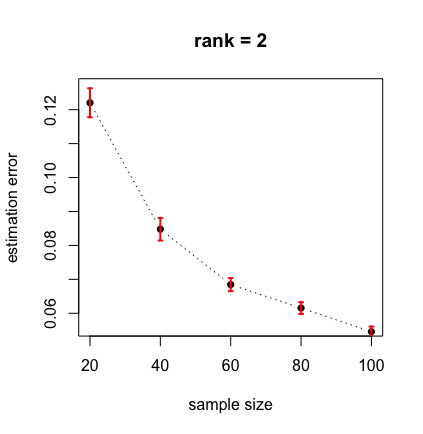}
\end{tabular}
\vspace{-0.2in}
\label{fig:spectral}
\end{figure}

Finally, we conduct a simulation to evaluate the empirical performance of the proposed spectral initialization Algorithms \ref{alg:initialization2} and \ref{alg:initialization3}. We simulate  the coefficient tensor $\cB^*\in \mathbb{R}^{30\times 20\times 10\times 5} = \sum_{k=1}^{r}w^*_k\bbeta^*_{k,1}\circ\bbeta^*_{k,2}\circ\bbeta^*_{k,3}\circ\bbeta^*_{k, 4}$. We generate the entries of $\bbeta^*_{k,j},\, k\in[2], j\in[3]$ from i.i.d.\ standard normal, and set $\bbeta^*_{k,4}$ as $(1,1,1,1,1)^{\top}$. We then normalize each vector to have a unit norm, and set $w_k^*=20$. We consider two ranks, $r=1$ and $r=2$, while we vary the sample size $n=\{20, 40, 60, 80, 100\}$. We then generate the error tensor $\cE_i$ with i.i.d.\ standard normal entries, and the response tensor $\cY_i\in \mathbb{R}^{30\times 20\times 10}$, with each entry missing with probability 0.5. For Algorithms \ref{alg:initialization2} and \ref{alg:initialization3},  we set $L = 30$, $\epsilon_{th} = 0.8$, and $\tau_{s_{j}}$ as $d_j$. Figure \ref{fig:spectral} reports the error, $\max_{k,j} \|\widehat{\bbeta}_{k,j}^{(0)}- \bbeta_{k,j}^*\|_2$, of the initial estimator based on 100 data replications. It is seen that, as the sample size increases, the estimation error decreases rapidly. This agrees with our finding in Proposition \ref{thm:initial}, and suggests that the constant initialization error bound in Assumptions \ref{ass:initialization} and \ref{ass:initialization_r} is to hold when $n$ is sufficiently large.

\section{Simulations}
\label{sec:simulations}

We carry out simulations to investigate the finite-sample performance of our proposed method. For easy reference, we call our method Partially ObServed dynamic Tensor rEsponse Regression (\texttt{POSTER}). We also compare with some alternative solutions. One competing method is the multiscale adaptive generalized estimating equations method (\texttt{MAGEE}) proposed by \cite{magee}, which integrated a voxel-wise approach with generalized estimating equations for adaptive analysis of dynamic tensor imaging data. Another competing method is the sparse tensor response regression method (\texttt{STORE}) proposed by \cite{store2017}, which considered a sparse tensor response regression model but did not incorporate fusion type smoothness constraint and can only handle completely observed data. In our analysis, \texttt{STORE} is applied to the complete samples only. Moreover, to examine the effect of utilizing the partially observed samples, and of incorporation of structural smoothness over time, we also compare to our own method but only applied to the completely observed samples, or without fusion constraint, which serve as two benchmarks.  
 
We consider two patterns for the unobserved entries, block missing in Section \ref{sec:simulation1} and random missing in Section \ref{sec:simulation2}. Both patterns are common in real data applications. For instance, in our neuroimaging example, individual subjects would miss some scheduled biannual scans, and as a result, the entire tensor images are unobserved, and the missing pattern is more likely a block missing. In our digital advertising example, on the other hand, some users may randomly react to only a subset of advertisements on certain days, and the missing pattern is closer to a random missing. Finally, in Section \ref{sec:simulation3}, we consider a model used in \cite{magee}. The data generation does not comply with our proposed model, and we examine the performance of our method under model misspecification. 

To evaluate the estimation accuracy, we report the estimation error of the coefficient tensor $\cB^*$ measured by $\|\widehat{\cB} - \cB^*\|_F$, and the estimation error of the decomposed components $\hat{\bbeta}_{k,j}$ measured by $\max_{k,j}\min\{\|\hat{\bbeta}_{k,j}-\bbeta^*_{k,j}\|,\|\hat{\bbeta}_{k,j}+\bbeta^*_{k,j}\|\}$. To evaluate the variable selection accuracy, we compute the true positive rate as the mean of $\textrm{TPR}_j$, and the false positive rate as the mean of $\textrm{FPR}_j$, where $\textrm{TPR}_j = K^{-1} \sum_{k=1}^K \sum_{l} \ind(\bbeta^*_{k,j,l} \ne 0, \widehat{\bbeta}_{k,j,l} \ne 0) / \sum_{l} \ind (\bbeta^*_{k,j,l} \ne 0)$ is the true positive rate of the estimator in mode $j$, and $\textrm{FPR}_j = K^{-1} \sum_{k=1}^K  \sum_{l} \ind( \bbeta_{k,j,l}^*  = 0,  \widehat{\bbeta}_{k,j,l}  \ne 0) / \sum_{l} \ind(\bbeta^*_{k,j,l} = 0)$ is the false positive rate of the estimator in mode $j$.

\subsection{Block missing}
\label{sec:simulation1}

In the first example, we simulate a fourth-order tensor response $\cY_i \in \mathbb{R}^{d_1\times d_2\times d_3\times T}$, where the fourth mode corresponds to the time dimension, and there are blocks of tensor entries missing along the time mode. More specifically, we generate the coefficient tensor $\cB^* \in \mathbb{R}^{d_1 \times d_2 \times d_3 \times T \times q}$ as $\cB^* = \sum_{k\in [r]} w_k^* \bbeta^*_{k,1} \circ \bbeta^*_{k,2} \circ \bbeta^*_{k,3}  \circ \bbeta^*_{k,4} \circ \bbeta^*_{k,5}$, where $d_1 = d_2 = d_3 = 32, T = 5, q = 5$, and the true rank $r=2$. We generate the entries of $\bbeta^*_{k,j}, j \in [4]$ as i.i.d.\ standard normal. We then apply the \texttt{Truncatefuse} operator on $\bbeta^*_{k,j}, j \in [3]$, with the true sparsity and fusion parameters $(s_{j},f_{j}), j \in [3]$, and apply the \texttt{Fuse} operator to $\bbeta^*_{k,4}$ with the true fusion parameter $f_{4}$. We set the true sparsity parameters $s_{j} = s_0 \times d_j, j \in [3]$ with $s_0=0.7$, and set the true fusion parameters $f_{j} = f_0 \times d_j, j \in [4]$, with $f_0 \in \{0.3, 0.7\}$. A smaller $f$ implies a smaller number of fusion groups in $\bbeta^*_{k,j}$. We set $\bbeta^*_{k,5}=(1,\ldots,1)^{\top}$, a vector of all ones. We then normalize each vector to have a unit norm. We set the weight $w^*_k \in \{30, 40\}$, with a larger weight indicating a stronger signal. Next, we generate the $q$-dimensional predictor vector $\xb_i$ whose entries are i.i.d.\ Bernoulli with probability 0.5, and the error tensor $\cE_i$, whose entries are i.i.d.\ standard normal. Finally, we generate the response tensor $\cY_i$ following model \eqref{eqn:true_model}. We set the blocks of entries of $\cY_i$ along the fourth mode randomly missing. Among all $n$ subjects, we set the proportion of subjects with missing values $m_n \in \{0.8,0.9\}$, and for each subject with missing values, we set the proportion of missing blocks along the time mode as $m_t \in \{0.4,0.6\}$. For example, $n=100$, $m_n = 0.8$ and $m_t = 0.4$ means there are 80 subjects out of 100 having partially observed tensors, and for each of those 80 subjects, the tensor observations at 2 out of 5 time points are missing. 

Table \ref{tab:our1} reports the average criteria based on 30 data replications with $m_n = 0.8$. The results with $m_n = 0.9$ are similar qualitatively and are reported in the Appendix. Since the method \texttt{MAGEE} of \cite{magee} does not decompose the coefficient tensor and does not carry out variable selection, the corresponding criteria of $\bbeta^*_{k,j}$ and selection are reported as NA. From this table, it is clearly seen that our proposed method outperforms all other competing methods in terms of both estimation accuracy and variable selection accuracy. 

\begin{table}[t!]
	\centering
	\footnotesize
	\caption{Simulation example with block missing, for varying missing proportions $m_n, m_t$, signal strength $w_k^*$, and fusion setting $f_0$. Reported are the average estimation errors of $\cB^*$ and $\bbeta^*_{k,j}$, and the true and false positive rates of selection based on 30 data replications (the standard errors in the parentheses). Five methods are compared: \texttt{STORE} of \citet{store2017}, \texttt{MAGEE} of  \citet{magee},  method applied to the complete data only (\texttt{Complete}), our method without the fusion constraint (\texttt{No-fusion}), and our proposed method (\texttt{POSTER}).}
\resizebox{\textwidth}{!}{
	\begin{tabular}{|c|c|c|c|cccc|} \hline
		$(m_n,m_t)$ & $w^*_{k}$ & $f_0$ & method &  Error of $\cB^*$ & Error of $\bbeta^*_{k,j}$ & TPR & FPR \\ \hline
		(0.8, 0.4) & 30 & 0.3 
		& \texttt{STORE} & 0.586 (0.055) & 0.992 (0.109) & 0.879 (0.016) & 0.369 (0.035) \\ 
		&  &  & \texttt{MAGEE} & 1.397 (0.005) & NA & NA & NA \\ 
		&  &  & \texttt{Complete}  &0.232 (0.051) & 0.366 (0.104) & 0.952 (0.017) & 0.104 (0.026) \\  	
		&  &  & \texttt{No-fusion} & 0.125 (0.003)& 0.112 (0.005) & 1.000 (0.000) & 0.120 (0.000) \\ 
		&  &  & \texttt{POSTER} & 0.069 (0.003) & 0.068 (0.005) & 1.000 (0.000) & 0.020 (0.004) \\ 
		\cline{3-8}
		&  & 0.7 
		& \texttt{STORE} & 0.574 (0.063) & 0.905 (0.113) & 0.878 (0.019) & 0.343 (0.043) \\ 
		&  &  & \texttt{MAGEE} & 1.411 (0.003) & NA & NA & NA \\ 
		&  &  & \texttt{Complete}  & 0.207 (0.038) & 0.259 (0.082) & 0.979 (0.008) & 0.103 (0.021) \\ 
		&  &  & \texttt{No-fusion} &0.120 (0.003) & 0.111 (0.006) & 1.000 (0.000) & 0.072 (0.000) \\ 
		& &  & \texttt{POSTER} & 0.102 (0.003) & 0.098 (0.006) & 1.000 (0.000) & 0.055 (0.003) \\
		\cline{2-8}
		& 40 & 0.3 & \texttt{STORE} & 0.287 (0.055) & 0.402 (0.104) & 0.957 (0.013) & 0.212 (0.028) \\
		&  &  & \texttt{MAGEE} & 1.233 (0.002) & NA & NA & NA \\ 
		&  & & \texttt{Complete} & 0.085 (0.022) & 0.087 (0.044) & 0.995 (0.005) & 0.036 (0.011) \\ 
		&  &  & \texttt{No-fusion}  & 0.115 (0.004) & 0.111 (0.005) & 1.000 (0.000) & 0.120 (0.000) \\ 
		&  &  & \texttt{POSTER} & 0.063 (0.004) & 0.067 (0.005) & 1.000 (0.000) & 0.020 (0.004) \\ 
		\cline{3-8}
		& & 0.7 & \texttt{STORE}  & 0.167 (0.036) & 0.160 (0.06) & 0.984 (0.009) & 0.131 (0.029) \\ 
		&  &  & \texttt{MAGEE} & 1.250 (0.002) & NA & NA & NA \\ 
		& &  &  \texttt{Complete} & 0.142 (0.030) & 0.190 (0.073) & 0.984 (0.008) & 0.107 (0.026) \\ 
		&  &  & \texttt{No-fusion} & 0.107 (0.003) & 0.115 (0.005) & 1.000 (0.000) & 0.093 (0.021) \\ 
		& &  & \texttt{POSTER} & 0.093 (0.004) & 0.094 (0.006) & 1.000 (0.000) & 0.074 (0.019) \\ 
		\cline{1-8}
		
		(0.8, 0.6) & 30 & 0.3 & \texttt{STORE} & 0.579 (0.057) & 0.975 (0.109) & 0.883 (0.016) & 0.360 (0.034) \\
		&  &  & \texttt{MAGEE} & 1.515 (0.004) & NA & NA & NA \\ 
		&  &  & \texttt{Complete}  & 0.233 (0.051)& 0.366 (0.104) & 0.952 (0.017) & 0.108 (0.026) \\ 
		& &  & \texttt{No-fusion}& 0.155 (0.006) & 0.146 (0.008) & 1.000 (0.000) & 0.120 (0.000) \\ 
		& &  & \texttt{POSTER} & 0.089 (0.006) & 0.091 (0.009) & 1.000 (0.000) & 0.023 (0.005) \\ 
		\cline{3-8}
		& & 0.7 
		& \texttt{STORE} & 0.434 (0.058) & 0.729 (0.120) & 0.924 (0.015) & 0.248 (0.034) \\ 
		&  &  & \texttt{MAGEE} & 1.528 (0.004) & NA & NA & NA \\ 
		& &  & \texttt{Complete} & 0.207 (0.038)& 0.259 (0.082) & 0.979 (0.008) & 0.103 (0.021) \\ 
		& &  & \texttt{No-fusion} & 0.151 (0.007) & 0.150 (0.009) & 1.000 (0.000) & 0.072 (0.000) \\ 
		&  &  & \texttt{POSTER} &0.128 (0.008)& 0.121 (0.010) & 1.000 (0.000) & 0.058 (0.002) \\ 
		\cline{2-8}
		& 40 & 0.3 & \texttt{STORE} & 0.228 (0.045) & 0.323 (0.096) & 0.971 (0.011) & 0.178 (0.021) \\ 
		&  &  & \texttt{MAGEE} & 1.310 (0.003) & NA & NA & NA \\  
		& &  & \texttt{Complete}& 0.090 (0.022) & 0.176 (0.073) & 0.983 (0.010) & 0.054 (0.016) \\ 
		& & & \texttt{No-fusion} & 0.142 (0.006) & 0.142 (0.008) & 0.999 (0.001) & 0.124 (0.003) \\ 
		& & & \texttt{POSTER} & 0.082 (0.006) & 0.089 (0.009) & 1.000 (0.000) & 0.023 (0.004) \\
		\cline{3-8}
		&  & 0.7 
		& \texttt{STORE}  & 0.228 (0.047) & 0.290 (0.090) & 0.969 (0.012) & 0.146 (0.029) \\ 
		&  &  & \texttt{MAGEE} & 1.325 (0.003) & NA & NA & NA \\ 
		& &  & \texttt{Complete} & 0.137 (0.022) & 0.205 (0.076) & 0.955 (0.016) & 0.159 (0.038) \\ 
		&  &  & \texttt{No-fusion}  & 0.131 (0.005) & 0.141 (0.010) & 0.999 (0.001) & 0.073 (0.002) \\ 
		&  &  & \texttt{POSTER} & 0.110 (0.006) & 0.122 (0.016) & 0.999 (0.001) & 0.061 (0.003) \\ \hline		
	\end{tabular}}
	\label{tab:our1}
\end{table}

The computational time of our method scales linearly with the sample size and tensor dimension. Consider the simulation setup with $m_n=0.8, m_t=0.4, w_{k}=30$, and $f_0 = 0.3$ as an example. When we fix $d_1=32$ and other parameters, the average computational time of our method was $112.5,  200.3, 384.2$ seconds for the sample size $n=100, 200, 300$, respectively. When we fix $n=100$ and other parameters, the average computational time of our method was $42.5,  82.3, 101.8$ seconds for the tensor dimension $d_1 = 10, 20, 30$, respectively. The reported computational time does not include tuning. All simulations were run on a personal computer with a 3.2 GHz Intel Core i5 processor.

\subsection{Random missing}
\label{sec:simulation2} 

In the second example, we simulate data similarly as in Section \ref{sec:simulation1}, but the entries of the response tensor are randomly missing. We set the observation probability $p \in \{0.3, 0.5\}$. For this setting, \texttt{MAGEE} cannot handle a tensor response with randomly missing entries, whereas \texttt{STORE} or our method applied to the complete data cannot handle either, since there is almost no complete $\cY_i$, with the probability of observing a complete $\cY_i$ being $p^{d_1 d_2 d_3 q}$. Therefore, we can only compare our proposed method with the variation that imposes no fusion constraint. Table \ref{tab:our11} reports the results based on 30 data replications. It is seen that incorporating the fusion structure clearly improves the estimation accuracy. Moreover, Table \ref{tab:our11} shows that the estimation error of our method decreases when the signal strength $w_k^*$ increases or when the observation probability $p$ increases. These patterns agree with our theoretical findings. 

\begin{table}[t!]
	\centering
	\scriptsize
	\caption{Simulation example with random missing, for varying observation probability $p$, signal strength $w^*_k$, and fusion setting $f_0$. Reported are the average estimation errors of $\cB^*$ and of $\bbeta^*_{k,j}$, and the true and false positive rates of selection based on 30 data replications (the standard errors in the parentheses). Two methods are compared: our method without the fusion constraint (\texttt{No-fusion}), and our proposed method (\texttt{POSTER}).}
	\resizebox{\textwidth}{!}{
	\begin{tabular}{|c|c|c|c|cccc|} \hline
		$p$& $w_{k}^*$ & $f_0$ & method & Error of $\cB^*$ & Error of $\bbeta^*_{k,j}$ & TPR & FPR \\ \hline
		0.5 & 30 & 0.3 & \texttt{No-fusion} & 0.091 (0.001) & 0.059 (0.001) & 1.000 (0.000) & 0.121 (0.001) \\ 
		&  &  & \texttt{POSTER} & 0.055 (0.001) & 0.037 (0.001) & 1.000 (0.000) & 0.021 (0.004) \\ 
		\cline{3-8}
		&  & 0.7 & \texttt{No-fusion} & 0.088 (0.001) & 0.056 (0.001) & 1.000 (0.000) & 0.099 (0.026) \\ 
		&  &  & \texttt{POSTER} & 0.079 (0.002) & 0.051 (0.001) & 1.000 (0.000) & 0.079 (0.024) \\ 		          \cline{2-8}
		& 40 & 0.3 & \texttt{No-fusion} & 0.068 (0.001) & 0.044 (0.001) & 1.000 (0.000) & 0.120 (0.000) \\ 
		&  &  & \texttt{POSTER} & 0.042 (0.001) & 0.029 (0.001) & 1.000 (0.000) & 0.019 (0.003) \\ 
		\cline{3-8}	
		&  & 0.7 & \texttt{No-fusion} & 0.066 (0.001) & 0.043 (0.001) & 1.000 (0.000) & 0.072 (0.000) \\ 
		&  & & \texttt{POSTER} & 0.059 (0.001) & 0.039 (0.001) & 1.000 (0.000) & 0.056 (0.003) \\ 		          \cline{1-8}
		0.3 & 30 & 0.3 & \texttt{No-fusion} & 0.119 (0.002) & 0.078 (0.002) & 0.998 (0.001) & 0.148 (0.023) \\ 
		& &  & \texttt{POSTER} & 0.077 (0.002) & 0.054 (0.002) & 1.000 (0.000) & 0.052 (0.016) \\ 
		\cline{3-8}  
		& & 0.7 & \texttt{No-fusion} & 0.113 (0.002) & 0.074 (0.002) & 0.998 (0.001) & 0.104 (0.026) \\ 
		&  &  & \texttt{POSTER} & 0.103 (0.002) & 0.066 (0.002) & 0.998 (0.001) & 0.086 (0.024) \\ 
		\cline{2-8}
		& 40 & 0.3 & \texttt{No-fusion} & 0.092 (0.020)& 0.060 (0.001) & 1.000 (0.000) & 0.120 (0.000) \\ 
		& &  & \texttt{POSTER} & 0.058 (0.001) & 0.042 (0.001) & 1.000 (0.000) & 0.025 (0.005) \\ 	
		\cline{3-8}  
		&  & 0.7 & \texttt{No-fusion} & 0.084 (0.001) & 0.054 (0.001) & 0.999 (0.000) & 0.074 (0.001) \\ 
		&  &  & \texttt{POSTER} & 0.075 (0.001) & 0.049 (0.001) & 1.000 (0.000) & 0.054 (0.030) \\ \hline	
		\end{tabular}}
	\label{tab:our11}
\end{table}

\subsection{Model misspecification}
\label{sec:simulation3}

In the third example, we simulate data from the model in \citet{magee}. Data generated this way does not comply with our proposed model \eqref{eqn:true_model}, and we examine the performance of our method under model misspecification. Following \citet{magee}, we simulate a third-order tensor response $\cY_i \in \mathbb{R}^{d_1\times d_2\times T}$, where the first two modes correspond to imaging space and the third mode corresponds to the time dimension, with $d_1 = d_2 = 88,\, T = 3$, and the sample size $n = 80$. At voxel $(j,k)$ the response of subject $i$ at time point $l$  is simulated according to
\begin{equation*} 
\cY_{i,j,k,l} = \xb^{\top}_{i,l} \bbeta_{j,k}^* + \epsilon_{i,j,k,l}, \quad i \in [n],  \; l \in [3].
\end{equation*}
The predictor vector $\xb_{i,l} = (1,x_{i,l,2},x_{i,l,3})^{\top}$, and we consider two settings of generating $\xb_{i,l}$. The first setting is that $x_{i,l,2}$ is time-dependent and is generated from a uniform distribution on $[l-1,l]$ for $l=1,2,3$, and $x_{i,l,3}$ is time independent and is generated from a Bernoulli distribution with probability 0.5. The second setting is that both $x_{i,l,2}$ and $x_{i,l,3}$ are time independent and are generated from a Bernoulli distribution with probability 0.5.  The error term $\bm{\epsilon}_{i,j,k} = (\epsilon_{i,j,k,1}, \epsilon_{i,j,k,2}, \epsilon_{i,j,k,3} )^{\top}$ is generated from a multivariate normal $N(0,\Sigma)$, where the diagonal entries of $\Sigma$ are 1 and $\mathrm{Corr}( \epsilon_{i,j,k,l_1}, \epsilon_{i,j,k,l_2} ) = 0.7^{|l_1-l_2|}$, $l_1, l_2 = 1,2,3$. The coefficient $\bbeta_{j,k}^* = ( 0,\beta_{j,k,2}^*, \beta_{j,k,3}^* )^{\top}$, and the coefficient image is divided into six different regions with two different shapes. Following \citet{magee}, we set $( \beta_{j,k,2}^*, \beta_{j,k,3}^* )$ to (0, 0), (0.05, 0.9), (0.1, 0.8), (0.2, 0.6), (0.3, 0.4) and (0.4, 0.2) in those six regions. Among the 80 subjects, the first half have their $88 \times 88$ images observed only at the first two time points.

\begin{figure}[t!]
\centering
\caption{True and estimated image of $\beta_{j,k,2}^*$. The top left panel is the true image of $\beta_{j,k,2}^*$ with six regions. The middle panels are the estimated images by \texttt{MAGEE}, and the right panels by our method \texttt{POSTER}. The top panels correspond to the time dependent covariates, and the bottom panels the time independent covariates. The estimation error (with the standard error in the parenthesis) based on 20 data replications is reported for each image.}
\vspace{0em}
\begin{tabular}{ccc}
\includegraphics[scale=0.38]{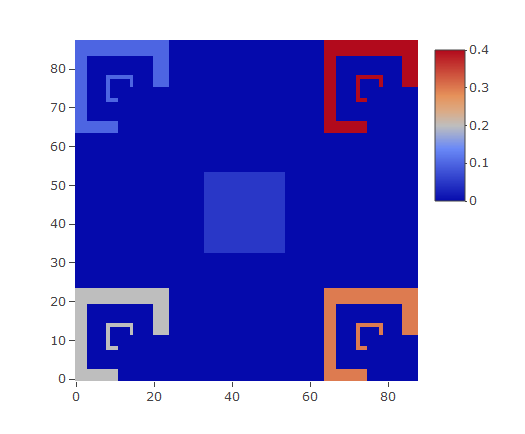} &
\includegraphics[scale=0.38]{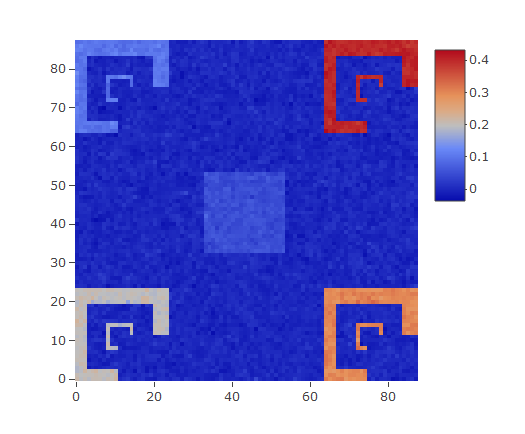} & 
\includegraphics[scale=0.35]{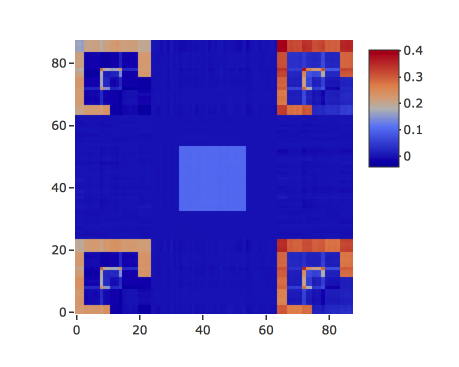} \\
(a) Truth & (b) \texttt{MAGEE}: 0.537 (0.003) & (c) \texttt{POSTER}: 0.468 (0.017) \\	
& 
\includegraphics[scale=0.38]{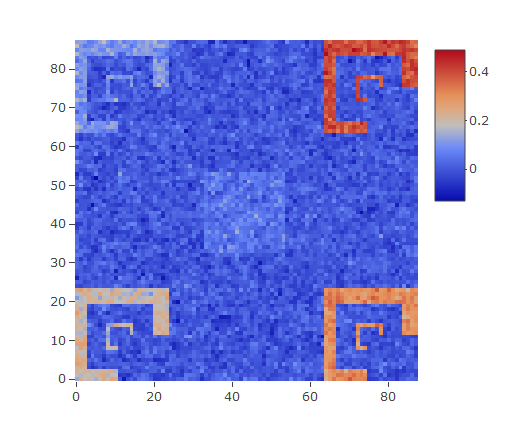} & 
\includegraphics[scale=0.36]{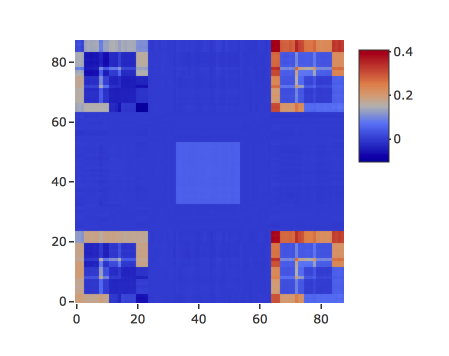} \\
 & (d) \texttt{MAGEE}: 0.682 (0.004) & (e) \texttt{POSTER}: 0.295 (0.009)
\end{tabular}
\label{fig:compare}
\end{figure}

Figure \ref{fig:compare} presents the true and estimated image of $\beta_{j,k,2}^*$, along with the estimation error of the coefficient tensor $\cB^*$. The standard error shown in parenthesis is calculated based on 20 replications. The results for $\beta_{j,k,3}^*$ are similar and hence are omitted. It is seen that our method is able to capture all six important regions in both settings of covariates, even if the model is misspecified. When the covariates are time dependent, our method is comparable to \citet{magee}. When the covariates are time independent, our estimator is more accurate compared to the method of \citet{magee}.

\section{Applications}
\label{sec:realdata}

We illustrate the proposed method with two real data applications. The first is a neuroimaging study, where about $50\%$ of  subjects have at least one imaging scan missing. The second is a digital advertising study, where about $95\%$ of tensor entries are missing.

\subsection{Neuroimaging application}
\label{sec:data1}

The first example is a neuroimaging study of dementia. Dementia is a broad category of brain disorders with symptoms associated with decline in memory and daily functioning \citep{alzheimer2012}. It is of keen scientific interest to understand how brain structures change and differ between dementia patients and healthy controls, which in turn would facilitate early disease diagnosis and development of effective treatment.

The data we analyze is from the Alzheimer's Disease Neuroimaging Initiative (ADNI, \url{http://adni.loni.usc.edu}), where anatomical MRI images were collected from $n=365$ participates every six months over a two-year period. Each MRI image, after preprocessing and mapping to a common registration space, is summarized in the form of a $32 \times 32 \times 32$ tensor. For each participant, there are at most five scans, but many subjects missed some scheduled scans, and 178 subjects out of 365 have at least one scan missing. For each subject, we stack the MRI brain images collected over time as a fourth-order tensor, which is to serve as the response $\cY_i$. Its dimension is $32 \times 32 \times 32 \times 5$, and there are block missing entries. Among these subjects, 127 have dementia and 238 are healthy controls. In addition, the baseline age and sex of the subjects were collected. As such, the predictor vector $\xb_i$ consists of the binary diagnosis status, age and sex. Our goal is to identify brain regions that differ between dementia patients and healthy controls, while controlling for other covariates.

\begin{figure}[t!]
\centering
\caption{Neuroimaging application example. Shown are the estimated coefficient tensor overlaid on a randomly selected brain image. Top to bottom: \texttt{MAGEE}, \texttt{STORE}, and our method \texttt{POSTER}. Left to right: frontal view, side view, and top view.}
\vspace{0em}

\includegraphics[scale=0.125]{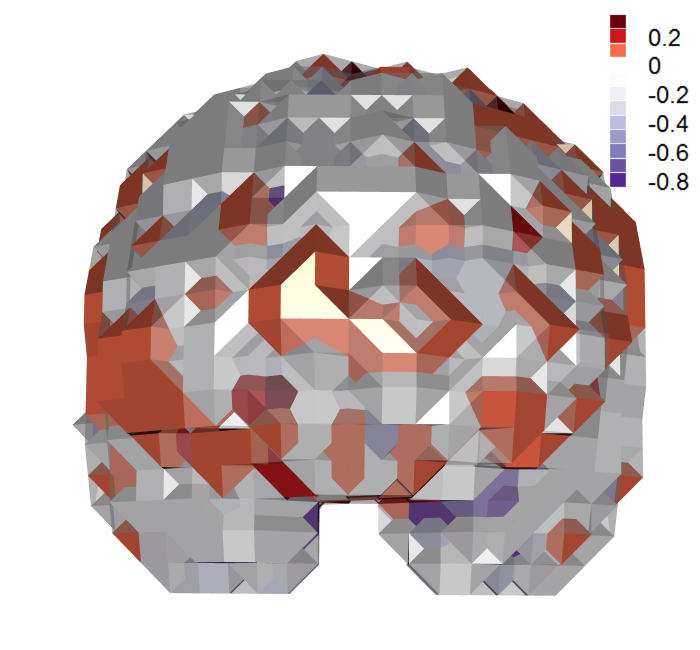}
\includegraphics[scale=0.125]{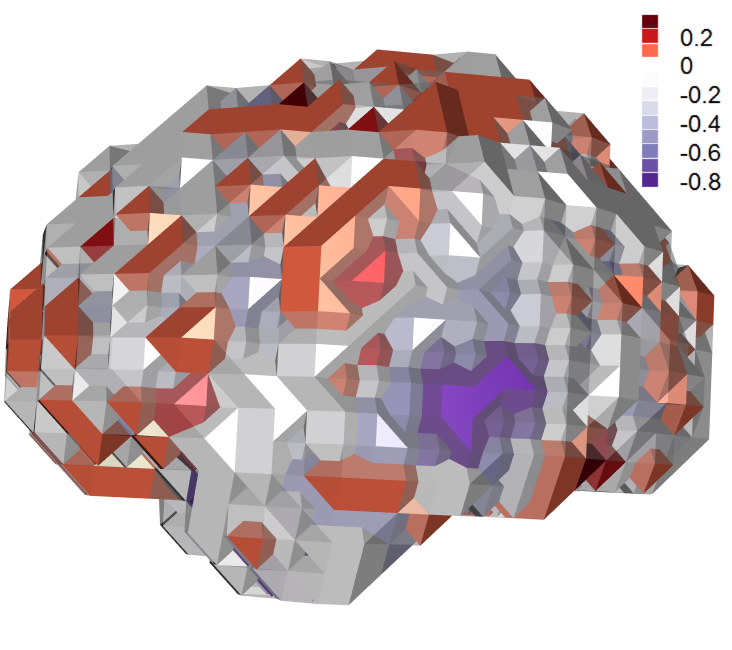}
\includegraphics[scale=0.125]{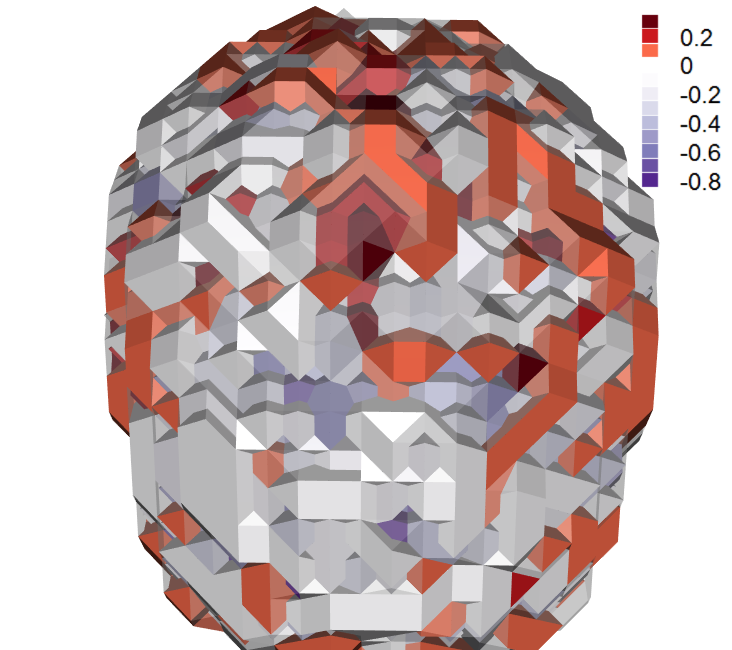}

\includegraphics[scale=0.13]{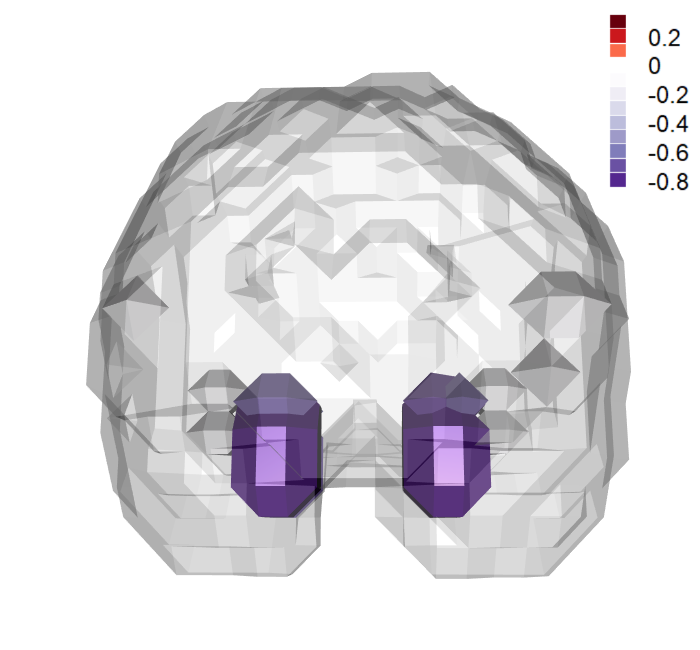}
\includegraphics[scale=0.13]{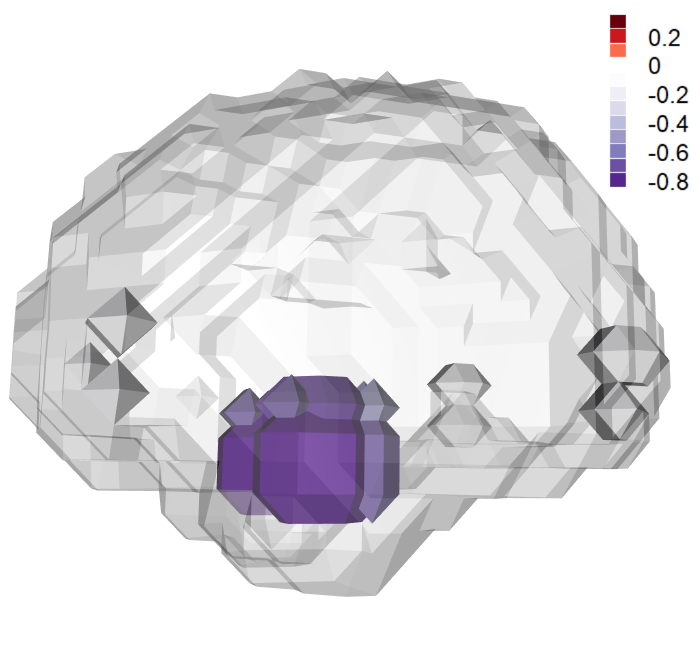}
\includegraphics[scale=0.13]{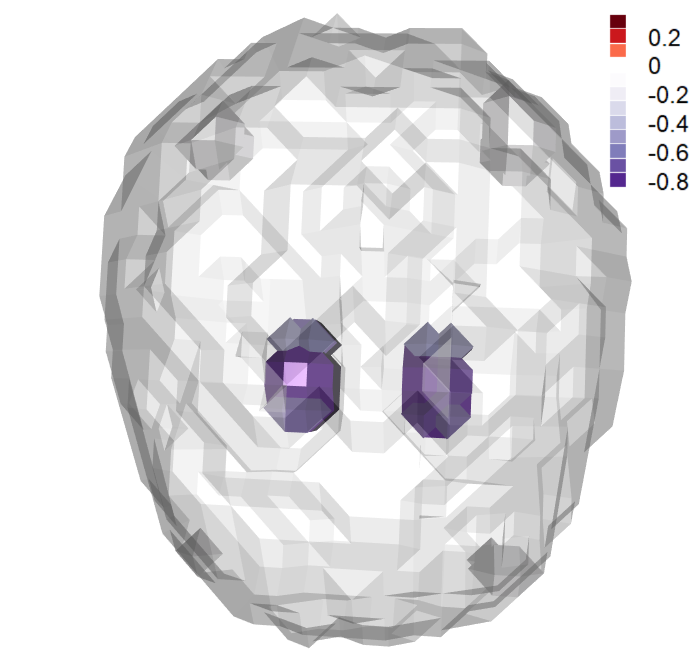}

\includegraphics[scale=0.13]{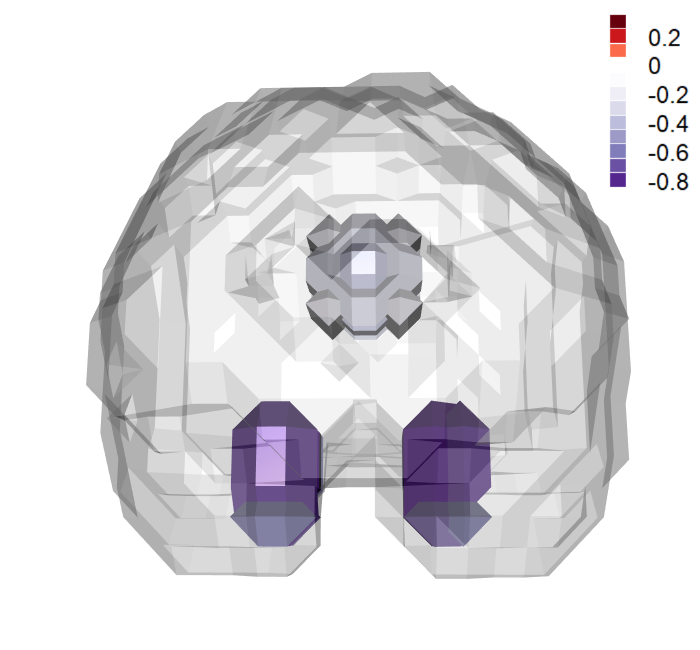}
\includegraphics[scale=0.13]{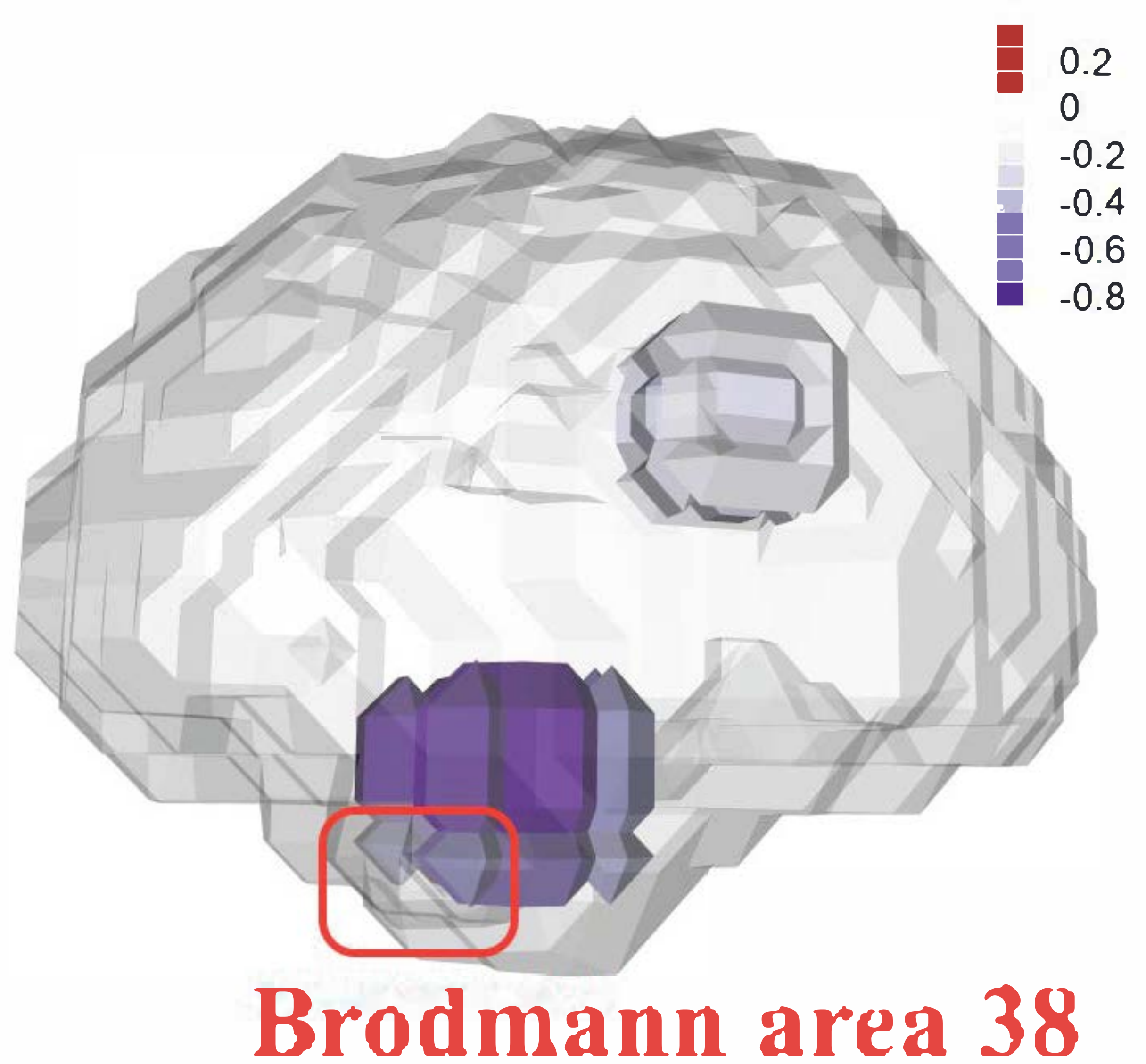}
\includegraphics[scale=0.13]{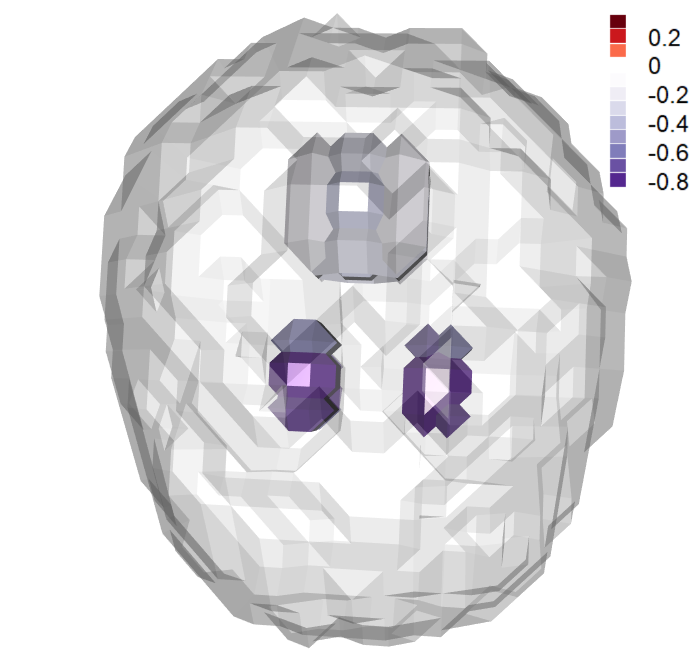}
\label{fig:compare1}
\vspace{0em}
\end{figure} 

We apply \texttt{MAGEE}, \texttt{STORE} and our \texttt{POSTER} method to this dataset. Figure \ref{fig:compare1} shows the heatmap of the estimated coefficient tensor 
at the baseline time point obtained by the three methods. It is seen that the estimate from \texttt{MAGEE} is noisy, which identifies a large number of regions with relatively small signals. Both \texttt{STORE} and \texttt{POSTER} identify several important brain regions, and the parameters in those identified regions are negative, indicating that those regions become less active for patients with dementia. The regions identified by the two methods largely agree with each other, with one exception, i.e., Brodmann area 38, which \texttt{POSTER} identifies but \texttt{STORE} does not. The regions identified by both include hippocampus and the surrounding medial temporal lobe. These findings are consistent with existing neuroscience literature. Hippocampus is found crucial in memory formation, and medial temporal lobe is important for memory storage \citep{smith2007}. Hippocampus is commonly recognized as one of the first regions in the brain to suffer damages for patients with dementia \citep{hampel2008}. There is also clear evidence showing that medial temporal lobe is damaged for dementia patients  \citep{visser2002}. In addition to those two important regions, our method also identifies a small part of the anterior temporal cortex, i.e., Brodmann area 38, which is highlighted in Figure \ref{fig:compare1}. This area is involved in language processing, emotion and memory, and is also among the first areas affect by Alzheimer's disease, which is the most common type of dementia \citep{delacourte1998}.

\subsection{Digital advertising application}
\label{sec:data2}

The second example is a digital advertising study of click-through rate (CTR) for some online advertising campaign. CTR is the number of times a user clicks on a specific advertisement divided by the number of times the advertisement is displayed. It is a crucial measure to evaluate the effectiveness of an advertisement campaign, and plays an important role in digital advertising pricing \citep{richardson2007predicting}.

The data we analyze is obtained from a major internet company over four weeks in May to June, 2016. The CTR of $n=80$ advertisement campaigns were recorded for 20 users by 2 different publishers. Since it is of more interest to understand the user behavior over different days of a week, the data were averaged by days of a week across the four-week period. For each campaign, we stack the CTR data of different users and publishers over seven days of the week as a third-order tensor, which is to serve as the response $\cY_i$. Its dimension is $20 \times 2 \times 7$, and there are 95\% entries missing. Such a missing percentage, however, is not uncommon in online advertising, since a user usually does not see every campaign by every publisher everyday. For each campaign, we also observe two covariates. One covariate is the topic of the advertisement campaign, which takes three categorical values, ``online dating", ``investment", or ``others". The other covariate is the total number of impressions of the advertisement campaign. The predictor vector $\xb_i$ consists of these two covariates. Our goal is to study how the topic and total  impression of an advertisement campaign affect its effectiveness measured by CTR.

\begin{figure}[t!]
\caption{Digital advertising application example. Shown are the estimated coefficient tensor. In each panel, the rows represent users and columns represent days of a week. The top panels are for the topic ``online dating", and the bottom panels for ``investment". The left panels are slices from the topic mode, and the right panels are slices from the impression mode.}
\centering
\includegraphics[scale=0.5]{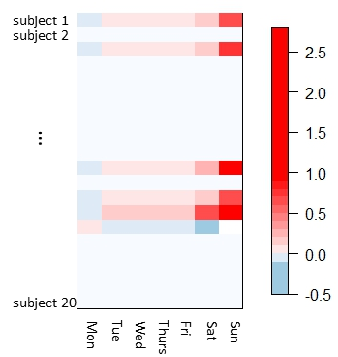}	
\includegraphics[scale=0.5]{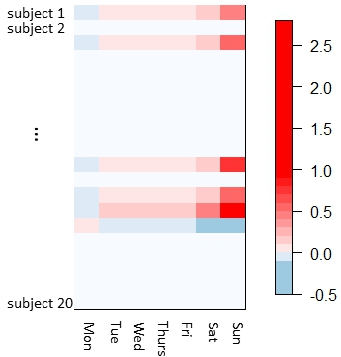}

\includegraphics[scale=0.5]{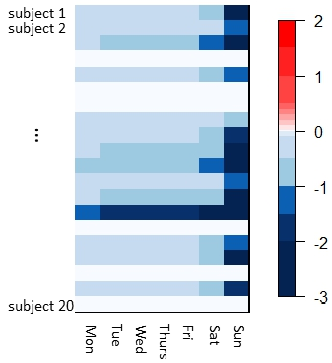}	
\includegraphics[scale=0.5]{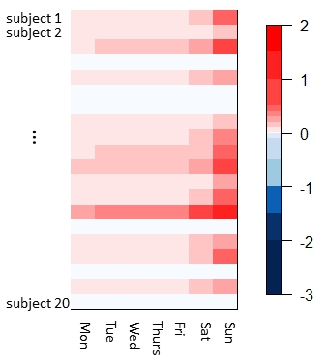}
\label{fig:ad}
\vspace{0em}
\end{figure}

Due to the large proportion of missing values and nearly random missing patterns, neither \texttt{MAGEE} nor \texttt{STORE} is applicable to this dataset. We apply our method. For the categorical covariate, topic, we create two dummy variables, one indicating whether the topic is ``online dating" or not, and the other indicating whether the topic is ``investment" or not. Figure \ref{fig:ad} shows the heatmap of the estimated coefficient tensor for one publisher, whereas the result for the other publisher is similar and is thus omitted. The rows of the heatmap represent the users and the columns represent the days of a week. We first consider the topic of ``online dating". The top left panel shows that, for this topic, the CTR is higher than other topics during the weekend. The top right panel shows that, if the total impression on ``online dating" increases, then the CTR increases more on weekends than weekdays.  It is also interesting to see that the topic of ``online dating" has a negative impact on the CTR on Mondays. We next consider the topic of ``investment". The bottom left panel shows that, for this topic, the CTR is lower than other topics for most users during the weekend. The bottom right panel shows that, if the total impression increases, the CTR increases more on weekends than weekdays. These findings are useful for managerial decisions. Based on the findings about ``online dating", one should increase the allocation of ``online dating" related advertisements on weekends, and decrease the allocation on Mondays. On the other hand, the allocation recommendation for ``investment" related advertisements are different. For most users, one should allocate more such advertisements during the early days of a week, and fewer during weekends. For a small group of users who seem to behave differently from the majority, some personalized recommendation regarding ``investment" advertisements can also be beneficial.

\baselineskip=17.5pt
\bibliographystyle{ims}
\bibliography{ref}

\begin{thebibliography}{51}
\expandafter\ifx\csname natexlab\endcsname\relax\def\natexlab#1{#1}\fi
\expandafter\ifx\csname url\endcsname\relax
  \def\url#1{\texttt{#1}}\fi
\expandafter\ifx\csname urlprefix\endcsname\relax\def\urlprefix{URL }\fi

\bibitem[{Anandkumar et~al.(2014)Anandkumar, Ge, Hsu and
  Kakade}]{anandkumar2014a}
\textsc{Anandkumar, A.}, \textsc{Ge, R.}, \textsc{Hsu, D.} and \textsc{Kakade,
  S.~M.} (2014).
\newblock A tensor approach to learning mixed membership community models.
\newblock \textit{Journal of Machine Learning Research} \textbf{15} 2239--2312.

\bibitem[{Bi et~al.(2018)Bi, Qu and Shen}]{bi2018multilayer}
\textsc{Bi, X.}, \textsc{Qu, A.} and \textsc{Shen, X.} (2018).
\newblock Multilayer tensor factorization with applications to recommender
  systems.
\newblock \textit{Annals of Statistics} \textbf{46} 3308--3333.

\bibitem[{Bruce et~al.(2017)Bruce, Murthi and Rao}]{bruce2017dynamic}
\textsc{Bruce, N.~I.}, \textsc{Murthi, B.} and \textsc{Rao, R.~C.} (2017).
\newblock A dynamic model for digital advertising: The effects of creative
  format, message content, and targeting on engagement.
\newblock \textit{Journal of marketing research} \textbf{54} 202--218.

\bibitem[{Bullmore and Sporns(2009)}]{bullmore2009complex}
\textsc{Bullmore, E.} and \textsc{Sporns, O.} (2009).
\newblock Complex brain networks: graph theoretical analysis of structural and
  functional systems.
\newblock \textit{Nature Reviews Neuroscience} \textbf{10} 186--198.

\bibitem[{Cai et~al.(2019)Cai, Li, Poor and Chen}]{clpc19}
\textsc{Cai, C.}, \textsc{Li, G.}, \textsc{Poor, H.} and \textsc{Chen, Y.}
  (2019).
\newblock Nonconvex low-rank tensor completion from noisy data.
\newblock \textit{NeurIPS} .

\bibitem[{Cai et~al.(2016)Cai, Li and Ma}]{cai2016optimal}
\textsc{Cai, T.~T.}, \textsc{Li, X.} and \textsc{Ma, Z.} (2016).
\newblock Optimal rates of convergence for noisy sparse phase retrieval via
  thresholded wirtinger flow.
\newblock \textit{The Annals of Statistics} \textbf{44} 2221--2251.

\bibitem[{Chen et~al.(2019)Chen, Raskutti and Yuan}]{ChenYuan2019}
\textsc{Chen, H.}, \textsc{Raskutti, G.} and \textsc{Yuan, M.} (2019).
\newblock Non-convex projected gradient descent for generalized low-rank tensor
  regression.
\newblock \textit{Journal of Machine Learning Research} \textbf{20} 1--37.

\bibitem[{Delacourte et~al.(1998)Delacourte, David et~al.}]{delacourte1998}
\textsc{Delacourte, A.}, \textsc{David, J.} \textsc{et~al.} (1998).
\newblock The biochemical pathway of neurofibrillary degeneration in aging and
  alzheimer's disease.
\newblock \textit{American Academy of Neurology} \textbf{52} 1158--1165.

\bibitem[{Feng et~al.(2019)Feng, Li, Song and Zhu}]{feng2019bayesian}
\textsc{Feng, X.}, \textsc{Li, T.}, \textsc{Song, X.} and \textsc{Zhu, H.}
  (2019).
\newblock Bayesian scalar on image regression with non-ignorable non-response.
\newblock \textit{Journal of the American Statistical Association} .

\bibitem[{Hampel et~al.(2008)Hampel, Burger et~al.}]{hampel2008}
\textsc{Hampel, H.}, \textsc{Burger, K.} \textsc{et~al.} (2008).
\newblock Core candidate neurochemical and imaging biomarkers of alzheimer's
  disease.
\newblock \textit{Alzheimer's and Dementia.} \textbf{4} 38--48.

\bibitem[{Han et~al.(2020)Han, Willett and Zhang}]{hwz20}
\textsc{Han, R.}, \textsc{Willett, R.} and \textsc{Zhang, A.} (2020).
\newblock An optimal statistical and computational framework for generalized
  tensor estimation.
\newblock \textit{arXiv} .

\bibitem[{Hao et~al.(2020)Hao, Zhang and Cheng}]{hzc20}
\textsc{Hao, B.}, \textsc{Zhang, A.} and \textsc{Cheng, G.} (2020).
\newblock Sparse and low-rank tensor estimation via cubic sketchings.
\newblock \textit{IEEE Transactions on Information Theory} \textbf{66}
  5927--5964.

\bibitem[{Jain and Oh(2014)}]{JO2014}
\textsc{Jain, P.} and \textsc{Oh, S.} (2014).
\newblock Provable tensor factorization with missing data.
\newblock \textit{Advances in Neural Information Processing Systems} \textbf{2}
  1431--1439.

\bibitem[{Kolda and Bader(2009)}]{kolda2009tensor}
\textsc{Kolda, T.~G.} and \textsc{Bader, B.~W.} (2009).
\newblock Tensor decompositions and applications.
\newblock \textit{SIAM review} \textbf{51} 455--500.

\bibitem[{Li and Zhang(2017)}]{li2016}
\textsc{Li, L.} and \textsc{Zhang, X.} (2017).
\newblock Parsimonious tensor response regression.
\newblock \textit{Journal of the American Statistical Association} \textbf{112}
  1131--1146.

\bibitem[{Li et~al.(2013)Li, Gilmore, Shen, Styner, Lin and Zhu}]{magee}
\textsc{Li, Y.}, \textsc{Gilmore, J.~H.}, \textsc{Shen, D.}, \textsc{Styner,
  M.}, \textsc{Lin, W.} and \textsc{Zhu, H.} (2013).
\newblock Multiscale adaptive generalized estimating equations for longitudinal
  neuroimaging data.
\newblock \textit{NeuroImage} \textbf{72} 91 -- 105.

\bibitem[{Ma(2013)}]{ma2013sparse}
\textsc{Ma, Z.} (2013).
\newblock Sparse principal component analysis and iterative thresholding.
\newblock \textit{Annals of Statistics} \textbf{41} 772--801.

\bibitem[{Madrid-Padilla and Scott(2017)}]{madrid2017}
\textsc{Madrid-Padilla, O.} and \textsc{Scott, J.} (2017).
\newblock Tensor decomposition with generalized lasso penalties.
\newblock \textit{Journal of Computational and Graphical Statistics}
  \textbf{26} 537--546.

\bibitem[{Rabusseau and Kadri(2016)}]{rabusseau2016}
\textsc{Rabusseau, G.} and \textsc{Kadri, H.} (2016).
\newblock Low-rank regression with tensor responses.
\newblock In \textit{Advances in Neural Information Processing Systems}.

\bibitem[{Richardson et~al.(2007)Richardson, Dominowska and
  Ragno}]{richardson2007predicting}
\textsc{Richardson, M.}, \textsc{Dominowska, E.} and \textsc{Ragno, R.} (2007).
\newblock Predicting clicks: estimating the click-through rate for new ads.
\newblock In \textit{Proceedings of the 16th international conference on World
  Wide Web}. ACM.

\bibitem[{Rinaldo(2009)}]{rinaldo2009}
\textsc{Rinaldo, A.} (2009).
\newblock Properties and refinements of the fused lasso.
\newblock \textit{The Annals of Statistics} \textbf{37} 2922--2952.

\bibitem[{Ryota and Taiji(2014)}]{ts14}
\textsc{Ryota, T.} and \textsc{Taiji, S.} (2014).
\newblock Spectral norm of random tensors.
\newblock \textit{arXiv} .

\bibitem[{Shen et~al.(2012)Shen, Pan and Zhu}]{shen2012}
\textsc{Shen, X.}, \textsc{Pan, W.} and \textsc{Zhu, Y.} (2012).
\newblock Likelihood-based selection and sharp parameter estimation.
\newblock \textit{Journal of American Statistical Association} \textbf{107}
  223--232.

\bibitem[{Smith and Kosslyn(2007)}]{smith2007}
\textsc{Smith, E.~E.} and \textsc{Kosslyn, S.~M.} (2007).
\newblock Cognitive psychology: Mind and brian.
\newblock \textit{New Jersey: Prentice Hall.} \textbf{21} 279--306.

\bibitem[{Sosa-Ortiz et~al.(2012)Sosa-Ortiz, Acosta-Castillo and
  Prince}]{alzheimer2012}
\textsc{Sosa-Ortiz, A.~L.}, \textsc{Acosta-Castillo, I.} and \textsc{Prince,
  M.~J.} (2012).
\newblock Epidemiology of dementias and alzheimer's disease.
\newblock \textit{Archives of medical research} \textbf{43} 600--608.

\bibitem[{Sun et~al.(2017)Sun, Lu, Liu and Cheng}]{sun2017provable}
\textsc{Sun, W.}, \textsc{Lu, J.}, \textsc{Liu, H.} and \textsc{Cheng, G.}
  (2017).
\newblock Provable sparse tensor decomposition.
\newblock \textit{Journal of the Royal Statistical Society, Series B}
  \textbf{79} 899--916.

\bibitem[{Sun and Li(2017)}]{store2017}
\textsc{Sun, W.~W.} and \textsc{Li, L.} (2017).
\newblock Store: Sparse tensor response regression and neuroimaging analysis.
\newblock \textit{Journal of Machine Learning Research} \textbf{18} 1--37.

\bibitem[{Sun and Li(2019)}]{SunLi2019}
\textsc{Sun, W.~W.} and \textsc{Li, L.} (2019).
\newblock Dynamic tensor clustering.
\newblock \textit{Journal of American Statistical Association} \textbf{114}
  1894 -- 1907.

\bibitem[{Tan et~al.(2018)Tan, Wang, Liu and Zhang}]{Tan2018sparse}
\textsc{Tan, K.~M.}, \textsc{Wang, Z.}, \textsc{Liu, H.} and \textsc{Zhang, T.}
  (2018).
\newblock Sparse generalized eigenvalue problem: optimalstatistical rates via
  truncated rayleigh flow.
\newblock \textit{Journal of the Royal Statistical Society: Series B}
  \textbf{80} 1057--1086.

\bibitem[{Tang et~al.(2019)Tang, Bi and Qu}]{tang2019individualized}
\textsc{Tang, X.}, \textsc{Bi, X.} and \textsc{Qu, A.} (2019).
\newblock Individualized multilayer tensor learning with an application in
  imaging analysis.
\newblock \textit{Journal of the American Statistical Association}  To Appear.

\bibitem[{Thung et~al.(2016)Thung, Wee, Yap and Shen}]{thung2016}
\textsc{Thung, K.-H.}, \textsc{Wee, C.-Y.}, \textsc{Yap, P.-T.} and
  \textsc{Shen, D.} (2016).
\newblock Identification of progressive mild cognitive impairment patients
  using incomplete longitudinal mri scans.
\newblock \textit{Brain Structure and Function} \textbf{221} 3979--3995.

\bibitem[{Tibshirani et~al.(2005)Tibshirani, Saunders, Rosset, Zhu and
  Knight}]{Tibshirani2005}
\textsc{Tibshirani, R.}, \textsc{Saunders, M.}, \textsc{Rosset, S.},
  \textsc{Zhu, J.} and \textsc{Knight, K.} (2005).
\newblock Sparsity and smoothness via the fused lasso.
\newblock \textit{Journal of the Royal Statistical Society Series B}
  \textbf{67} 91--108.

\bibitem[{Visser et~al.(2002)Visser, Verhey, Hofman et~al.}]{visser2002}
\textsc{Visser, P.}, \textsc{Verhey, F.}, \textsc{Hofman, P.} \textsc{et~al.}
  (2002).
\newblock Medial temporal lobe atrophy predicts alzheimer's disease in patients
  with minor cognitive impairment.
\newblock \textit{Journal of Neurology, Neurosurgery and Psychiatry.}
  \textbf{72} 491--497.

\bibitem[{Vounou et~al.(2010)Vounou, Nichols, Montana, Initiative
  et~al.}]{vounou2010discovering}
\textsc{Vounou, M.}, \textsc{Nichols, T.~E.}, \textsc{Montana, G.},
  \textsc{Initiative, A. D.~N.} \textsc{et~al.} (2010).
\newblock Discovering genetic associations with high-dimensional neuroimaging
  phenotypes: A sparse reduced-rank regression approach.
\newblock \textit{Neuroimage} \textbf{53} 1147--1159.

\bibitem[{Wang and Li(2020)}]{WL20}
\textsc{Wang, M.} and \textsc{Li, L.} (2020).
\newblock Learning from binary multiway data: Probabilistic tensor
  decomposition and its statistical optimality.
\newblock \textit{Journal of Machine Learning Research} \textbf{21} 1--38.

\bibitem[{Wang and Zhu(2017)}]{WangZhu2016}
\textsc{Wang, X.} and \textsc{Zhu, H.} (2017).
\newblock Generalized scalar-on-image regression models via total variation.
\newblock \textit{Journal of the American Statistical Association} \textbf{112}
  1156--1168.

\bibitem[{Wang et~al.(2016)Wang, Sharpnack, Smola and Tibshirani}]{wang2016}
\textsc{Wang, Y.}, \textsc{Sharpnack, J.}, \textsc{Smola, A.} and
  \textsc{Tibshirani, R.} (2016).
\newblock Trend filtering on graphs.
\newblock \textit{Journal of Machine Learning Research} \textbf{17} 1--41.

\bibitem[{Wang et~al.(2015{\natexlab{a}})Wang, Tung, Smola and
  Anandkumar}]{wang2015b}
\textsc{Wang, Y.}, \textsc{Tung, H.-Y.}, \textsc{Smola, A.} and
  \textsc{Anandkumar, A.} (2015{\natexlab{a}}).
\newblock Fast and guaranteed tensor decomposition via sketching.
\newblock \textit{Advances in Neural Information Processing Systems} .

\bibitem[{Wang et~al.(2015{\natexlab{b}})Wang, Gu, Ning and Liu}]{wang2015high}
\textsc{Wang, Z.}, \textsc{Gu, Q.}, \textsc{Ning, Y.} and \textsc{Liu, H.}
  (2015{\natexlab{b}}).
\newblock High dimensional em algorithm: statistical optimization and
  asymptotic normality.
\newblock \textit{NeurIPS} \textbf{28}.

\bibitem[{Xia and Yuan(2017)}]{dong2017}
\textsc{Xia, D.} and \textsc{Yuan, M.} (2017).
\newblock On polynomial time methods for exact low rank tensor completion.
\newblock \textit{Foundations of Computational Mathematics}  1--49.

\bibitem[{Xia et~al.(2020)Xia, Yuan and Zhang}]{xyz20}
\textsc{Xia, D.}, \textsc{Yuan, M.} and \textsc{Zhang, C.} (2020).
\newblock Statistically optimal and computationally efficient low rank tensor
  completion from noisy entries.
\newblock \textit{Annals of Statistics} .

\bibitem[{Xu et~al.(2019)Xu, Hu and Wang}]{xhw19}
\textsc{Xu, Z.}, \textsc{Hu, J.} and \textsc{Wang, M.} (2019).
\newblock Generalized tensor regression with covariates on multiple modes.
\newblock \textit{arXiv} .

\bibitem[{Xue and Qu(2020)}]{XueQu2019}
\textsc{Xue, F.} and \textsc{Qu, A.} (2020).
\newblock Integrating multisource block-wise missing data in model selection.
\newblock \textit{Journal of the American Statistical Association} \textbf{0}
  1--14.

\bibitem[{Yin et~al.(2015)Yin, Cui, Chen, Hu and Zhou}]{yin2015dynamic}
\textsc{Yin, H.}, \textsc{Cui, B.}, \textsc{Chen, L.}, \textsc{Hu, Z.} and
  \textsc{Zhou, X.} (2015).
\newblock Dynamic user modeling in social media systems.
\newblock \textit{ACM Transactions on Information Systems} \textbf{33} 1--44.

\bibitem[{Yuan and Zhang(2016)}]{yuan2016on}
\textsc{Yuan, M.} and \textsc{Zhang, C.} (2016).
\newblock On tensor completion via nuclear norm minimization.
\newblock \textit{Foundations of Computational Mathematics} \textbf{16}
  1031--1068.

\bibitem[{Yuan and Zhang(2017)}]{yuan2017incoherent}
\textsc{Yuan, M.} and \textsc{Zhang, C.} (2017).
\newblock Incoherent tensor norms and their applications in higher order tensor
  completion.
\newblock \textit{IEEE Transactions on Information Theory} \textbf{63}
  6753--6766.

\bibitem[{Yuan and Zhang(2013)}]{yuan2013}
\textsc{Yuan, X.-T.} and \textsc{Zhang, T.} (2013).
\newblock Truncated power method for sparse eigenvalue problems.
\newblock \textit{Journal of Machine Learning Research} \textbf{14} 899--925.

\bibitem[{Zhang(2019)}]{zhang2019cross}
\textsc{Zhang, A.} (2019).
\newblock Cross: Efficient low-rank tensor completion.
\newblock \textit{Annals of Statistics} \textbf{47} 936--964.

\bibitem[{Zhang et~al.(2019)Zhang, Allen, Zhu and Dunson}]{zhang2019tensor}
\textsc{Zhang, Z.}, \textsc{Allen, G.~I.}, \textsc{Zhu, H.} and \textsc{Dunson,
  D.} (2019).
\newblock Tensor network factorizations: Relationships between brain structural
  connectomes and traits.
\newblock \textit{NeuroImage} \textbf{197} 330--343.

\bibitem[{Zhou et~al.(2013)Zhou, Li and Zhu}]{zhou2013}
\textsc{Zhou, H.}, \textsc{Li, L.} and \textsc{Zhu, H.} (2013).
\newblock Tensor regression with applications in neuroimaging data analysis.
\newblock \textit{Journal of the American Statistical Association} \textbf{108}
  540--552.

\bibitem[{Zhu et~al.(2014)Zhu, Shen and Pan}]{Zhu2014}
\textsc{Zhu, Y.}, \textsc{Shen, X.} and \textsc{Pan, W.} (2014).
\newblock Structural pursuit over multiple undirected graphs.
\newblock \textit{Journal of the American Statistical Association} \textbf{109}
  1683--1696.

\end{thebibliography}

\end{document}